\pdfoutput=1

\documentclass[11pt]{article}

\usepackage[final]{acl}

\usepackage{times}
\usepackage{latexsym}

\usepackage[T1]{fontenc}

\usepackage[utf8]{inputenc}

\usepackage{microtype}

\usepackage{inconsolata}

\usepackage{enumitem}
\usepackage{etoolbox}
\usepackage{booktabs,makecell,tabularx} 
\usepackage{resizegather}
\usepackage{multirow}
\usepackage{amssymb}
\usepackage{amsmath}
\usepackage{graphicx}
\usepackage{tcolorbox}
\usepackage{subcaption}
\usepackage{colortbl}
\usepackage{listings}
\usepackage{xcolor}
\usepackage{hypcap}
\usepackage{caption}

\title{UnifiedVisual: A Framework for Constructing Unified Vision-Language Datasets}

\author{Pengyu Wang\thanks{{} {} Equal contribution. Order is random.},
Shaojun Zhou\footnotemark[1], 
Chenkun Tan, 
Xinghao Wang,
\\
{\bf Wei Huang},
{\bf Zhen Ye},
{\bf Zhaowei Li},
{\bf Botian Jiang},
{\bf Dong Zhang},
{\bf Xipeng Qiu\thanks{{} {} Corresponding author.}} \\
\textsuperscript{1} Fudan University \\
\texttt{\{pywang22, sjzhou24, cktan25\}@m.fudan.edu.cn} \\
\texttt{\{xpqiu\}@fudan.edu.cn} \\}

\begin{document}
\maketitle
\begin{abstract}
Unified vision large language models (VLLMs) have recently achieved impressive advancements in both multimodal understanding and generation, powering applications such as visual question answering and text-guided image synthesis. However, progress in unified VLLMs remains constrained by the lack of datasets that fully exploit the synergistic potential between these two core abilities. Existing datasets typically address understanding and generation in isolation, thereby limiting the performance of unified VLLMs. To bridge this critical gap, we introduce a novel dataset construction framework, \textbf{UnifiedVisual}, and present \textbf{UnifiedVisual-240K}, a high-quality dataset meticulously designed to facilitate mutual enhancement between multimodal understanding and generation. UnifiedVisual-240K seamlessly integrates diverse visual and textual inputs and outputs, enabling comprehensive cross-modal reasoning and precise text-to-image alignment. Our dataset encompasses a wide spectrum of tasks and data sources, ensuring rich diversity and addressing key shortcomings of prior resources. Extensive experiments demonstrate that models trained on UnifiedVisual-240K consistently achieve strong performance across a wide range of tasks. Notably, these models exhibit significant mutual reinforcement between multimodal understanding and generation, further validating the effectiveness of our framework and dataset. We believe UnifiedVisual represents a new growth point for advancing unified VLLMs and unlocking their full potential.
\footnote{Our code and datasets will be available at \url{https://github.com/fnlp-vision/UnifiedVisual}.}
\end{abstract}

\section{Introduction}

\begin{figure}[!t]
  \centering
  \includegraphics[width=0.95\columnwidth]{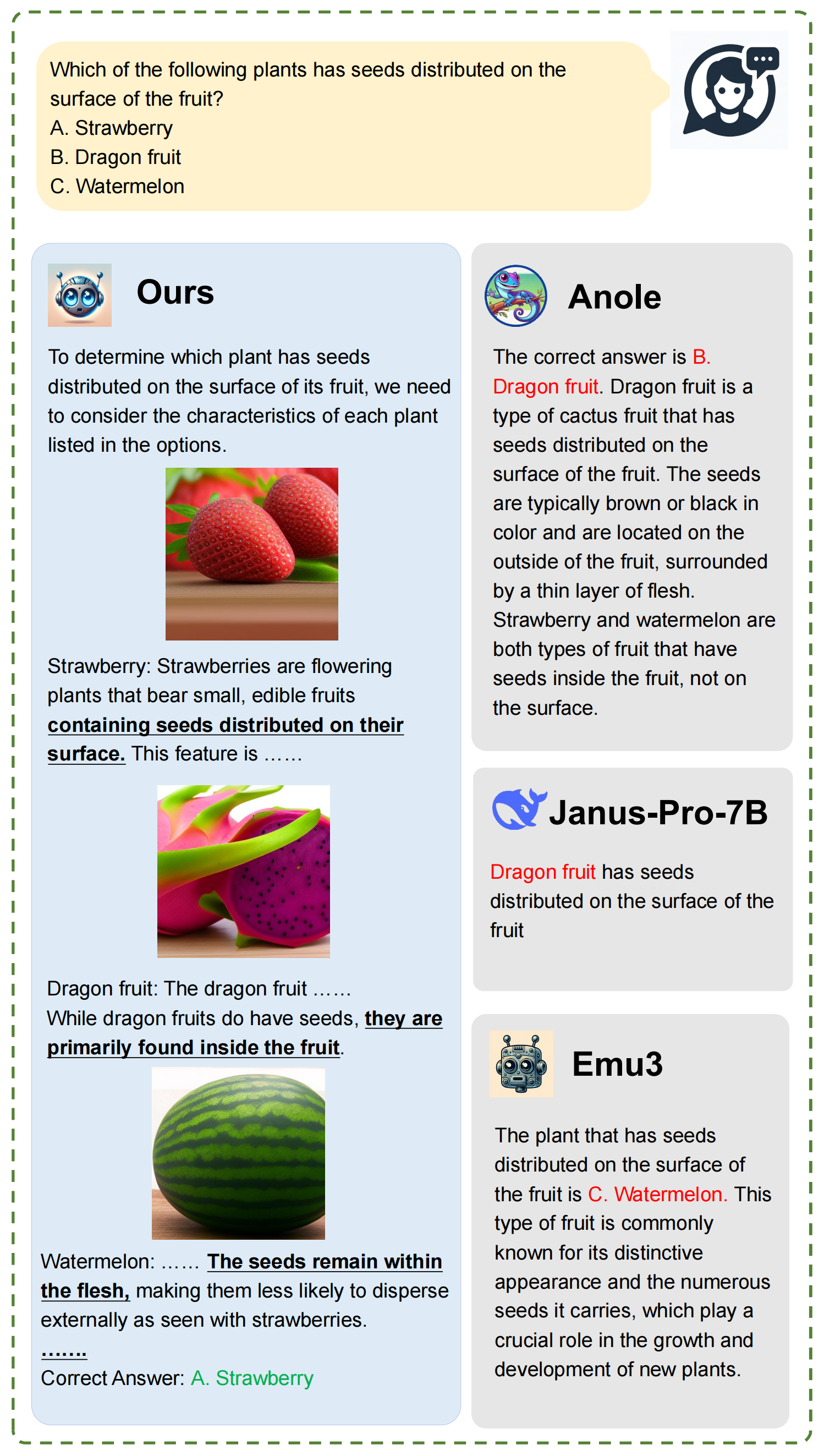}
  \caption{As illustrated in this figure, in contrast to other unified VLLMs that fail to answer even simple questions correctly due to conflicts between multimodal understanding and generation, our model trained on {UnifiedVisual-240K} is able to construct  multimodal reasoning chains and provide accurate answers.}
  \label{fig: main_1} 
  \vspace{-10pt}
\end{figure}

Vision large language models (VLLMs) have made significant progress in visual understanding, evolving from basic image captioning to complex visual inferences \cite{liu2024visual,dai2023instructblip}. Currently, there is growing interest in unified models capable of both multimodal understanding and generation. These models aim to integrate multimodal understanding and generation capabilities, enabling them to handle a variety of tasks such as image captioning, visual question answering, and image generation \cite{team2024chameleon,wu2024janus,tong2024metamorph}. Unified VLLMs have gained widespread attention due to their ability to combine multimodal understanding and generation in a single model. This unification not only simplifies the deployment and application process but also provides the potential for mutual enhancement between generative and discriminative capabilities. As a result, this area of research is becoming an increasingly prominent development field.

However, the development of powerful unified VLLMs hinges on access to high-quality training datasets.
While several existing datasets have facilitated progress, they fall short of fully unlocking the synergistic potential between multimodal understanding and generation.
Ideally, a unified VLLM should achieve substantial improvements by leveraging the interaction between these two capabilities. Yet, in practice, models trained on current datasets often exhibit limited integration, failing to achieve effective mutual reinforcement between understanding and generation \cite{wang2024emu3,wang2024illume}. This highlights a critical limitation in the design and quality of existing datasets, which are unable to fully stimulate the desired synergy.

To address these challenges, we propose a novel dataset construction framework, \textbf{UnifiedVisual}, and introduce \textbf{UnifiedVisual-240K}, a new dataset designed to enhance the interaction between multimodal understanding and generation. UnifiedVisual-240K incorporates the following key features: 
\textit{First}, the instructions may include both visual and textual information, encouraging holistic integration of multimodal context for accurate responses. 
\textit{Second}, the responses may also consist of both visual and textual elements, requiring the model to excel in both textual reasoning and multimodal generation. This duality ensures that textual reasoning guides precise image generation, while the generated images, in turn, enhance textual reasoning. This mutual reinforcement between the two modalities enables the model to achieve superior performance. 
\textit{Finally}, UnifiedVisual-240K exhibits significant diversity in both task types and data sources, effectively promoting the interaction between understanding and generative capabilities.

To validate the effectiveness of UnifiedVisual-240K, we trained unified VLLM models using this dataset. Experimental results show that models trained on UnifiedVisual-240K consistently outperform those trained on existing datasets across a wide range of tasks. Notably, we observed substantial mutual enhancement between the models’ understanding and generative capabilities, fully demonstrating the advantages of our dataset.

In summary, our contributions are as follows:
\begin{itemize}[itemsep=1pt, leftmargin=10pt, parsep=0pt, topsep=1pt]
\item 
We propose UnifiedVisual, a unified vision-language dataset construction framework that prioritizes the synergistic interaction between understanding and generative capabilities while ensuring task and data source diversity.
\item 
We construct UnifiedVisual-240K, a high-quality dataset tailored for unified VLLMs.
\item 
Experimental results demonstrate that models trained on UnifiedVisual-240K achieve superior performance and exhibit mutual enhancement between multimodal understanding and generation.
\end{itemize}


\section{Related Work}

\paragraph{Unified Visual Understanding and Generation.}
In recent years, research on unifying image understanding and generation within a single visual large language model (VLLM) has garnered significant attention. 
Early studies primarily achieved image generation by integrating image generation models (e.g., diffusion models) on top of large language models (LLMs) \cite{sun2023generative,wu2023next,li2024mini,ge2024seed}. More recently, \citet{tong2024metamorph} demonstrated remarkable results by connecting LLMs and diffusion models through a simple projection layer.
Inspired by the success of LLMs in next-step prediction tasks, recent studies have explored representing and generating images in a fully autoregressive manner using discrete visual tokens \cite{yu2023scaling,chen2023pixart,wang2024emu3,liu2024world,chern2024anole}. To achieve high performance in both image understanding and generation, some research efforts have proposed decoupling these two tasks. For instance, Transfusion \cite{zhou2024transfusion} and Show-o \cite{xie2024show} employ autoregressive text modeling for image understanding tasks while adopting visual diffusion modeling to accomplish image generation. In contrast, Janus \cite{wu2024janus} introduces two distinct image representations, specifically designed to address the differing granularity requirements of image understanding and generation.
Overall, exploration of unified VLLM architectures continues to progress.

\begin{figure}[!t]
  \centering
  \includegraphics[width=0.8\columnwidth]{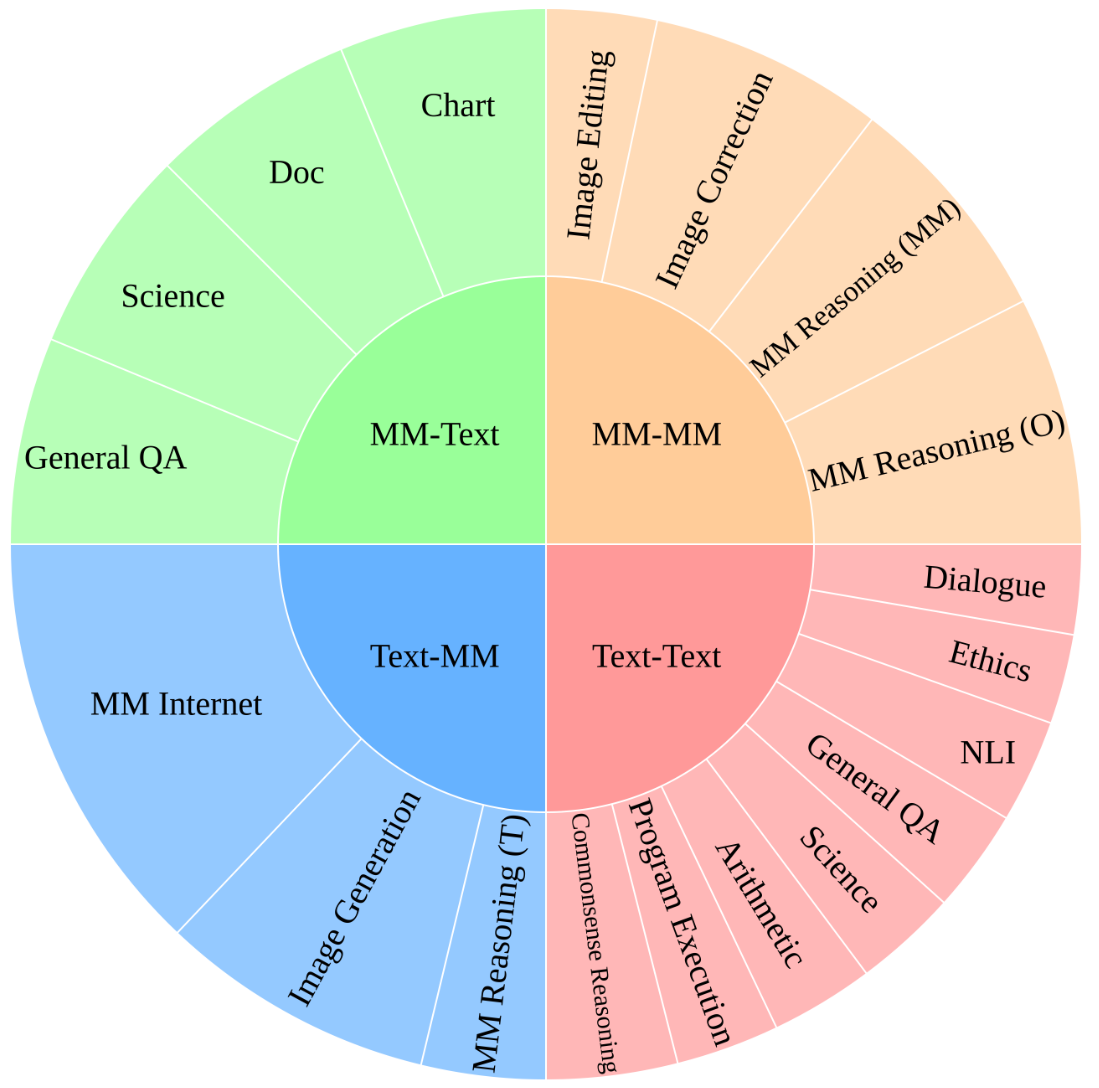}
  \caption{The proportions of different sub-datasets in UnifiedVisual-240K. The innermost layer of the chart represents the "\textit{input type} - \textit{output type}", such as Text-MM, which indicates that these datasets feature textual input and multimodal output.}
  \label{fig: UnifiedVisual-240Kset} 
  \vspace{-10pt}
\end{figure}

\paragraph{Training Datasets for Unified VLLM.}
\label{sec: related_work_training_dataset}
Given the unique characteristics of unified VLLMs, we divided the training dataset into four major categories, as shown in Figure \ref{fig: UnifiedVisual-240Kset}. Among them, datasets for pure text generation
are both abundant and of high quality \cite{shao2024case2code,li2024llava,zhang2024metaalign}. In contrast, datasets for multimodal generation are relatively narrow in scope and limited in scale. 
Currently, the most widely used multimodal generation datasets 
mainly cover image generation and image editing. However, these tasks themselves are relatively straightforward, as they typically involve direct mappings from instructions to images or simple image modifications
\cite{deng2009imagenet,brooks2023instructpix2pix,fu2023guiding,qu2024tokenflow}. Additionally, there exist interleaved image-text datasets crawled from the internet, but the association between images and text in such datasets is often weak \cite{zhu2024multimodal,laurenccon2024obelics}.

The scarcity of multimodal generation datasets not only limits the application of models in related downstream tasks but also introduces potential conflicts between multimodal understanding and generation during training. These conflicts make it challenging to achieve mutual enhancement of the two capabilities, potentially impacting the model's performance on complex tasks. To address these challenges, we propose a unified vision-language dataset construction framework to overcome the current limitations in training datasets. 

\section{Methodology}
\label{sec: methodology}

In this section, we first provide a detailed introduction to our vision-language dataset construction framework, \textbf{UnifiedVisual}. Following that, we introduce \textbf{UnifiedVisual-240K}, a dataset constructed following the UnifiedVisual framework.

\subsection{UnifiedVisual}

As discussed in Section \ref{sec: related_work_training_dataset}, the training datasets for unified VLLM can be categorized into two types: \textbf{understanding datasets} that only contain pure text outputs, and \textbf{generation datasets} that involve multimodal generation. 
Given the abundance and high quality of existing understanding datasets, we can directly leverage these well-established resources.
In contrast, generation datasets are typically narrower in scope and more limited in scale. To address this, UnifiedVisual introduces a novel and comprehensive framework for constructing generation datasets. Specifically, we focus on three key aspects to construct a more diverse and comprehensive generation dataset: (1) Visual Generation, (2) Multimodal Reasoning, and (3) Multimodal Internet Dataset. In the sections that follow, we will discuss each construction method in detail. 
For better clarity, we also provide illustrations in Appendix \ref{sec: appendix_dataset_construction_illustration}, and the complete prompt templates can be found in Appendix \ref{sec:appendix_prompt_template}.

\subsubsection{Visual Generation}

Visual Generation encompasses Image Generation, Image Editing, and Image Correction. 
Unlike existing datasets that primarily focus on generating or editing images based on simple descriptions or instructions, our goal is to integrate visual understanding and textual reasoning to tackle more complex visual generation challenges.

\paragraph{Image Generation} 
Image generation involves generating images that correspond to textual descriptions, serving as a foundational task in training unified VLLMs. However, existing image generation datasets often emphasize direct mappings between textual elements and images, which limits their ability to handle more intricate generation requirements. To address this, we propose two enhanced approaches:

\textit{Topic- and Scene-Based Generation}: (1) We propose several topics and corresponding scenes, then generate image captions that implicitly, rather than explicitly, describe the desired image content. 
(2) We use embedding models to filter similar captions, ensuring data diversity.
(3) We use GPT-4 to generate a reasoning process (rationale) explaining the content and details to be generated, followed by DALL-E-3 for image synthesis. 


\textit{Category- and Image-Based Generation}: (1) We collect a diverse set of authentic images, removing duplicates. (2) Based on the images and their categories, we use GPT-4 to generate instructions that describe image characteristics and related information, again focusing on implicit rather than explicit descriptions. (3) We then use GPT-4 to generate a detailed rationale based on the caption and category information, outlining the logic behind the desired image. The final data point consists of the caption, rationale, and the original image.


\paragraph{Image Editing} Existing image editing datasets typically consist of simple pairs of images and editing instructions that require straightforward modifications. However, these basic datasets may not effectively enhance a model's capacity to comprehend and execute sophisticated visual generation instructions. To address this limitation, we enhance existing image editing data through a two-step approach: (1) We transform simple editing instructions into more nuanced prompts that necessitate deeper understanding and planning. (2) We use GPT-4o to analyze these enhanced instructions and generate reasoning rationales outlining the editing objectives and intended outcomes.

\paragraph{Image Correction} To further enhance the model's capability in capturing fine-grained image details, we introduced a more sophisticated task paradigm: image correction. This task requires the model to evaluate image-description consistency and, when discrepancies are identified, analyze the inconsistencies before regenerating an image that fully aligns with the given description. We implement this through a three-stage process: (1) We modify existing image captions to create descriptions that maintain the core theme while introducing controlled variations in specific visual elements. (2) We utilize StableDiffusion to generate images containing intentional discrepancies based on these modified descriptions. (3) We employ GPT-4o to systematically analyze the generated images against the original descriptions, automatically identifying inconsistencies and providing detailed modification rationales. The final data point includes the original caption, the generated image, the analysis rationale, and the original image.

\subsubsection{Multimodal Reasoning}


Multimodal Reasoning focuses on the synergistic interplay between multimodal understanding and generation. During the reasoning process, multimodal reasoning drives the generation of necessary visual content, while the generated visual content, as part of the reasoning rationale, in turn facilitates better multimodal understanding. This design emulates human reasoning processes, where individuals often combine textual thinking with visual aids (such as mental sketches or imagined scenes) to collaboratively solve complex problems.

\paragraph{MM Reasoning (O)} 
In multimodal tasks, answering questions often requires 
careful attention to specific details within the \textbf{\underline{o}riginal input images}.
Following \citet{shao2024visual}, we construct questions that demand reasoning rationale incorporating snapshots of critical details from the original image.


\paragraph{MM Reasoning (MM)} To enhance the model's multimodal reasoning capabilities, we construct data points that require joint reasoning across both image and text modalities. The dataset construction process is as follows:
(1) We collect a diverse set of images and use the CLIP model to remove duplicates.
(2) GPT-4o is employed to generate reasoning questions based on the collected images. These questions are designed to require reasoning processes that integrate both visual and textual content. Questions that fail to meet this criterion are discarded.
(3) The input image and the generated question are provided to GPT-4o, which produces a rationale. When necessary, textual descriptions are used in place of images.
(4) The textual descriptions from step 3 are rewritten into keywords using GPT-4. These keywords are then used to retrieve images from tools like Bing Search, ensuring stylistic consistency across the images used in the questions and rationales.
(5) CLIP similarity scores are computed between the descriptions generated in step 3 and the retrieved images. Only the images with the highest similarity scores are retained.
Because \textbf{the input is \underline{m}ulti\underline{m}odal}, we refer to this construction method as MM Reasoning (MM).

\paragraph{MM Reasoning (T)} 
Beyond multimodal input scenarios, we also design multimodal reasoning task based on \textbf{purely \underline{t}extual inputs}. The process is as follows: 
(1) GPT-4 is used to generate text-only questions that require reasoning aided by generated images.
(2) The generated questions are deduplicated using embedding model to ensure diversity and uniqueness.
(3) GPT-4 is then tasked with answering these questions, providing detailed rationales while replacing image-related components with textual descriptions when necessary.
(4) DALL-E-3 is used to generate images based on the image descriptions generated in step 3.

\subsubsection{Multimodal Internet Dataset}

To further enhance the diversity and naturalness of the dataset, we process and transform interleaved text-image data sourced from the internet.

\paragraph{MM Internet} 
We construct this dataset based on a large collection of diverse multimodal data crawled from the internet. To improve data quality, we draw inspiration from \citet{chen2024comm} and design a multi-perspective filtering strategy. This strategy leverages pre-trained VLLMs to ensure coherence and semantic consistency between sentences and their associated images. Furthermore, we generate questions for these multimodal data, ensuring that the answers align precisely with the corresponding text-image data.

\subsubsection{ Compared to Existing Practice}
UnifiedVisual introduces several key innovations over existing datasets. \textit{First}, its synergistic design uniquely integrates multimodal understanding and generation, embedding complex reasoning into generation tasks and incorporating multimodal generation within reasoning chains for mutual reinforcement. \textit{Second}, UnifiedVisual significantly broadens task diversity, covering visual generation, multimodal reasoning, and large-scale internet-sourced data, thus overcoming the limited scope of prior datasets. \textit{Third}, by using indirect prompts and constructing reasoning chains that interleave text and images, UnifiedVisual enables models to learn more advanced reasoning processes, rather than simply mapping inputs to direct responses as in traditional datasets. \textit{Finally}, the framework is highly extensible, enabling efficient and large-scale dataset construction to support future unified VLLM development.

\subsection{UnifiedVisual-240K}

Using the above methods, we ultimately constructed 120k \textbf{generation samples}. The sources and final quantities of each type of data are shown in Table \ref{tab:UnifiedVisual-240K-generation}. 
Additionally, we sampled 60K data points from LLaVA-CoT \cite{xu2024llava} and CoT-Collection \cite{kim2023cot}, respectively, to create our \textbf{understanding samples}. Together with the generation samples, these form our UnifiedVisual-240K.
Its composition and distribution are illustrated in Figure \ref{fig: UnifiedVisual-240Kset}. 
More details about the dataset construction can be found in Appendix \ref{sec: appendix_dataset_construction}. Additionally, we provide examples for each subset of UnifiedVisual-240K in Appendix \ref{sec:appendix_example_data}.

\section{Experimental Setup}

\subsection{Unified VLLM}

In this section, we select Anole \cite{chern2024anole} as the base model for training and evaluation. Among all open-source unified VLLMs, Anole stands out as a representative model built on the transformer architecture. It adopts a unified processing approach for various modalities and supports multimodal outputs that can include any number of images. These capabilities make Anole particularly suitable as the base model for our experiments. Specifically, Anole represents images as discrete tokens. After generating these image tokens, the image decoder converts the discrete visual tokens back into images. For more details regarding the training procedure and inference, please refer to Appendix \ref{sec:appendix_exp_setup}.

\begin{table*}[t]
    \centering
    \resizebox{0.8\textwidth}{!}{
    \begin{tabular}{ l l l l l l l l}
    \toprule
    
    \textbf{Model}                       & \textbf{RWQA} & \textbf{MMVP} & \textbf{SQA} & \textbf{VStar} & \textbf{MME} & \textbf{POPE} & \textbf{Avg.}\\
    \midrule
    Anole                       & 32.0 & 10.0 & 46.7 & 15.7 & 841.4 & 65.8 & 33.4 \\
    Anole-NormalData            & \underline{37.9} & 7.3  & 53.4 & \underline{30.9} & 952.9 & \underline{75.9} & 39.9\\
    Anole-UnifiedVisual$_{T}$ & \underline{37.9} & \underline{20.0} & 55.2 & 29.8 & \underline{1316.5} & 72.1 & \underline{43.7} \\
    Anole-UnifiedVisual$_{MM}$ & 36.1 & 14.7 & \underline{55.3} & 28.3 & 1125.3 & 70.6 & 40.9 \\
    \rowcolor{gray!20}
    Anole-UnifiedVisual         & \textbf{39.7} & \textbf{24.0} & \textbf{56.2} & \textbf{33.0} & \textbf{1371.2} & \textbf{76.1} & \textbf{46.3}\\
      
    \bottomrule
    \end{tabular}
    }
    \caption{This table presents the results of the multimodal understanding evaluation. The best results are highlighted in \textbf{bold}, while the second-best results are marked with an \underline{underline} for clarity.}
    \label{tab:mm-understanding-results}
\end{table*}

\subsection{Evaluation and Metrics}


\paragraph{Multimodal Understanding} To evaluate multimodal understanding capabilities, we conduct evaluations on six widely-used benchmarks: RealworldQA \cite{realworldqa}, MMVP \cite{tong2024eyes}, ScienceQA \cite{lu2022learn}, VStar \cite{wu2023vguidedvisualsearch}, MME \cite{fu2024mmecomprehensiveevaluationbenchmark}, and POPE \cite{li2023evaluating}. For RealworldQA, MMVP, ScienceQA, and VStar, accuracy is used as the evaluation metric. GPT-4 is employed to determine whether the model's output match the ground truth, and accuracy is then calculated. Notably, for MMVP, a response is only considered correct if both paired questions are answered correctly. For MME and POPE, we first use GPT-4 to summarize the model's output as either "yes" or "no" and then use the official repository's code to compute the final metrics. Specifically, for MME, we report the total score for MME Perception and MME Cognition. For POPE, we report its F1 score.

\paragraph{Multimodal Generation} To evaluate visual generation capabilities, we use MS-COCO \cite{lin2014microsoft} and GenEval \cite{ghosh2024geneval}. For MS-COCO, we report the CLIP score as our evaluation metric. For GenEval, we use the official evaluation code\footnote{\url{https://github.com/djghosh13/geneval}} for assessment and report the overall score.

\paragraph{Textual Reasoning} To assess the model's pure text reasoning ability, we use AlpacaEval \cite{alpaca_eval}. Following the official AlpacaEval\footnote{\url{https://github.com/tatsu-lab/alpaca_eval}}, we use GPT-4 for evaluation. A higher win rate indicates greater helpfulness of the response.

\subsection{Experimental Details} 
During training, we utilized 64 NVIDIA H100 80G GPUs, set the batch size to 512, and the maximum sequence length to 4096. We used the AdamW optimizer with a 5\% warm-up step and the cosine decay learning rate scheduler. The model was trained for 2 epochs with a maximum learning rate of 2e-5. 
For inference, we used greedy decoding to generate textural responses for reproducibility. 
After decoding, we used Anole's vision decoder to transform the generated vision tokens into images.

\section{Experiments}

\subsection{Baselines}


\paragraph{Anole-NormalData} Following prior works \cite{ma2024janusflow,li2024synergen}, we trained Anole using a combination of textual understanding data, multimodal understanding data, and multimodal generation data. Specifically, the understanding data is identical to that of UnifiedVisual-240K, while the multimodal generation data was derived from an equivalent amount of Laion\footnote{\url{https://huggingface.co/datasets/dclure/laion-aesthetics-12m-umap}} \cite{schuhmann2022laion}. 
Laion is a high-quality dataset carefully filtered by high aesthetic scores, making it a popular choice for training advanced image generation models \cite{xie2024show}.
This data was subsequently transformed into the instruction-following format as outlined by \citet{tong2024metamorph}.

\paragraph{Anole-UnifiedVisual$_{T}$} To investigate the interaction between multimodal understanding and generation within UnifiedVisual-240K, we introduced an additional baseline model trained exclusively on the understanding subset of UnifiedVisual-240K.

\paragraph{Anole-UnifiedVisual$_{MM}$} Similarly, we added another baseline model trained solely on the generation subset of UnifiedVisual-240K.

\label{sec: main_results_mm_generation}
\begin{figure}[t]
  \centering
  \includegraphics[width=.85\columnwidth]{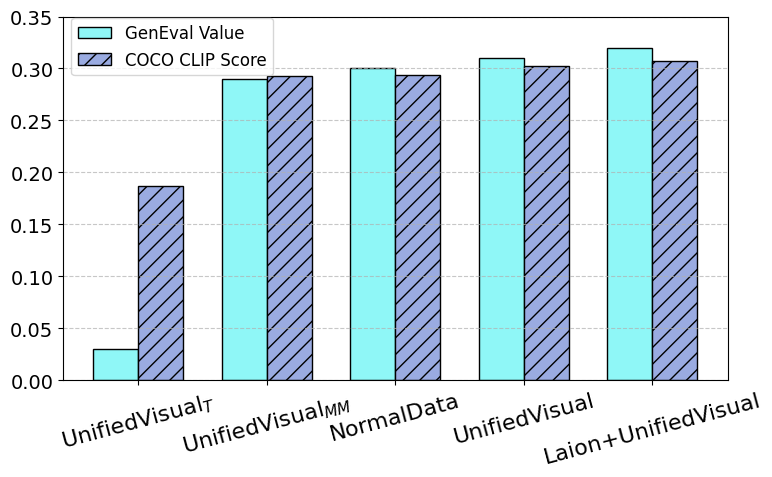}
  \caption{Generation evaluation for different models.}
  \label{fig:geneval-overall-score}
\end{figure}

\begin{figure*}[h]
    \centering
    \begin{subfigure}{.32\linewidth}
        \centering
        \includegraphics[width=\linewidth]{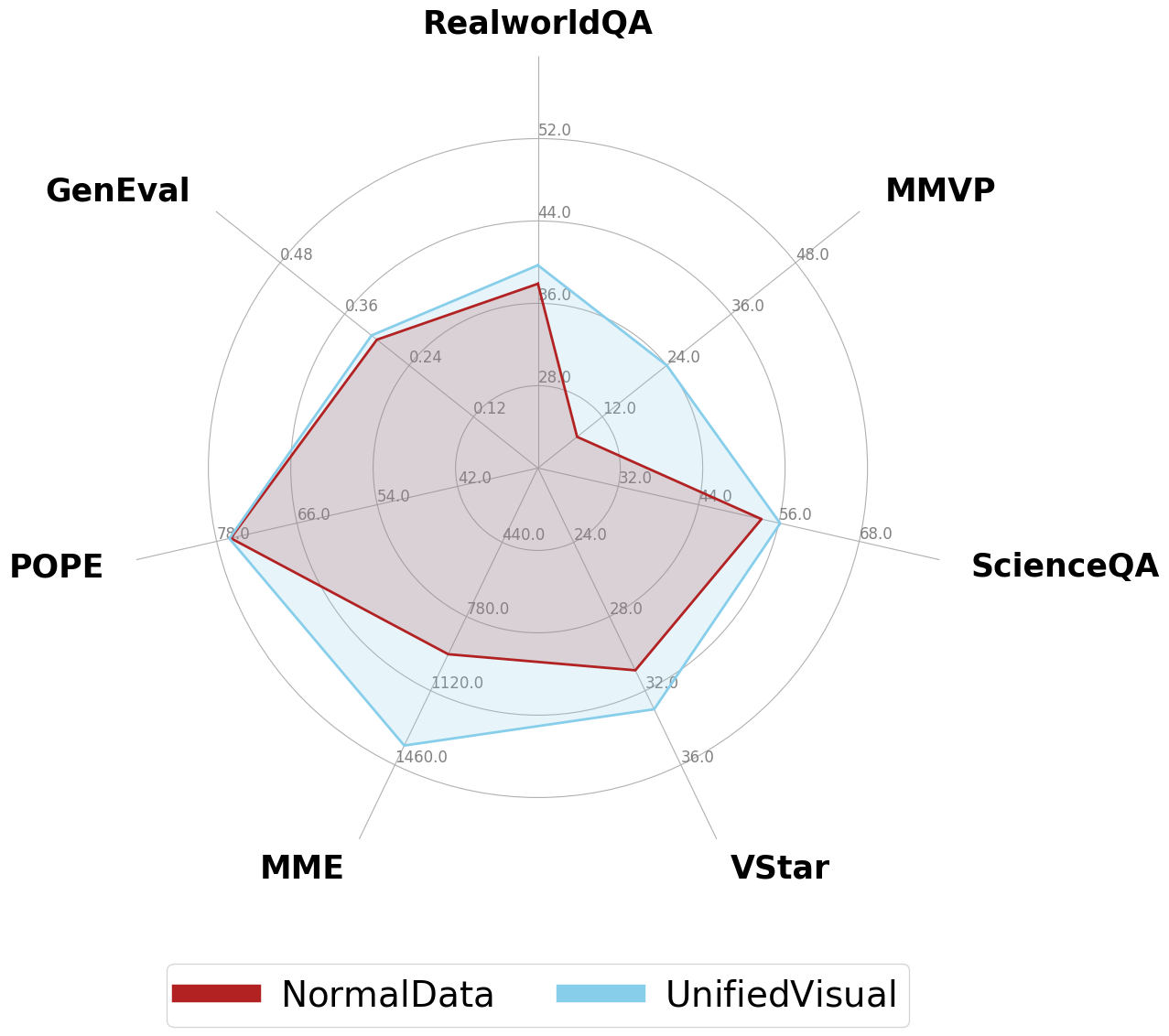}
        \caption{Anole}
    \end{subfigure}
    \hfill
    \begin{subfigure}{.32\linewidth}
        \centering
        \includegraphics[width=\linewidth]{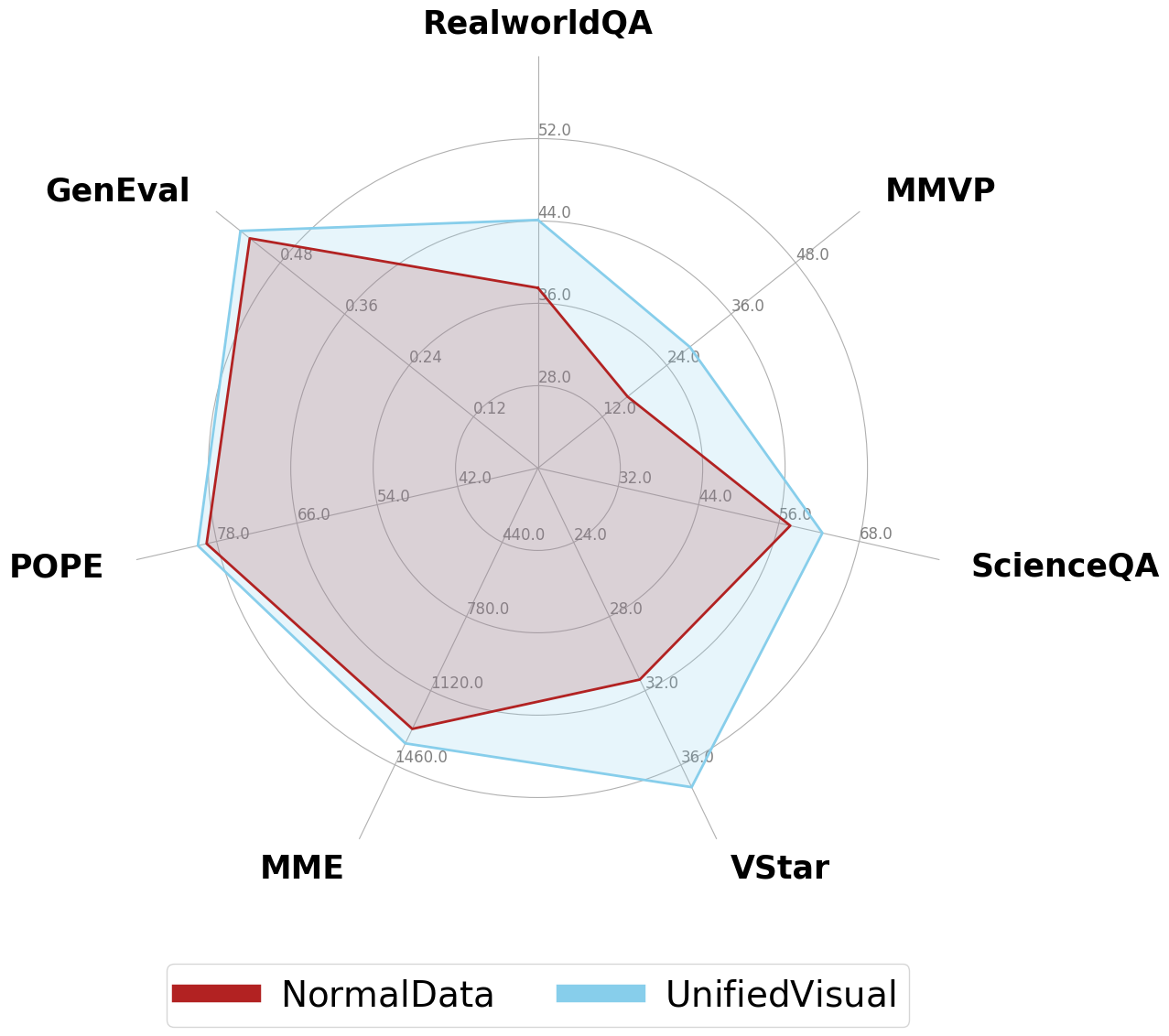}
        \caption{Liquid}
    \end{subfigure}
    \hfill
    \begin{subfigure}{.32\linewidth}
        \centering
        \includegraphics[width=\linewidth]{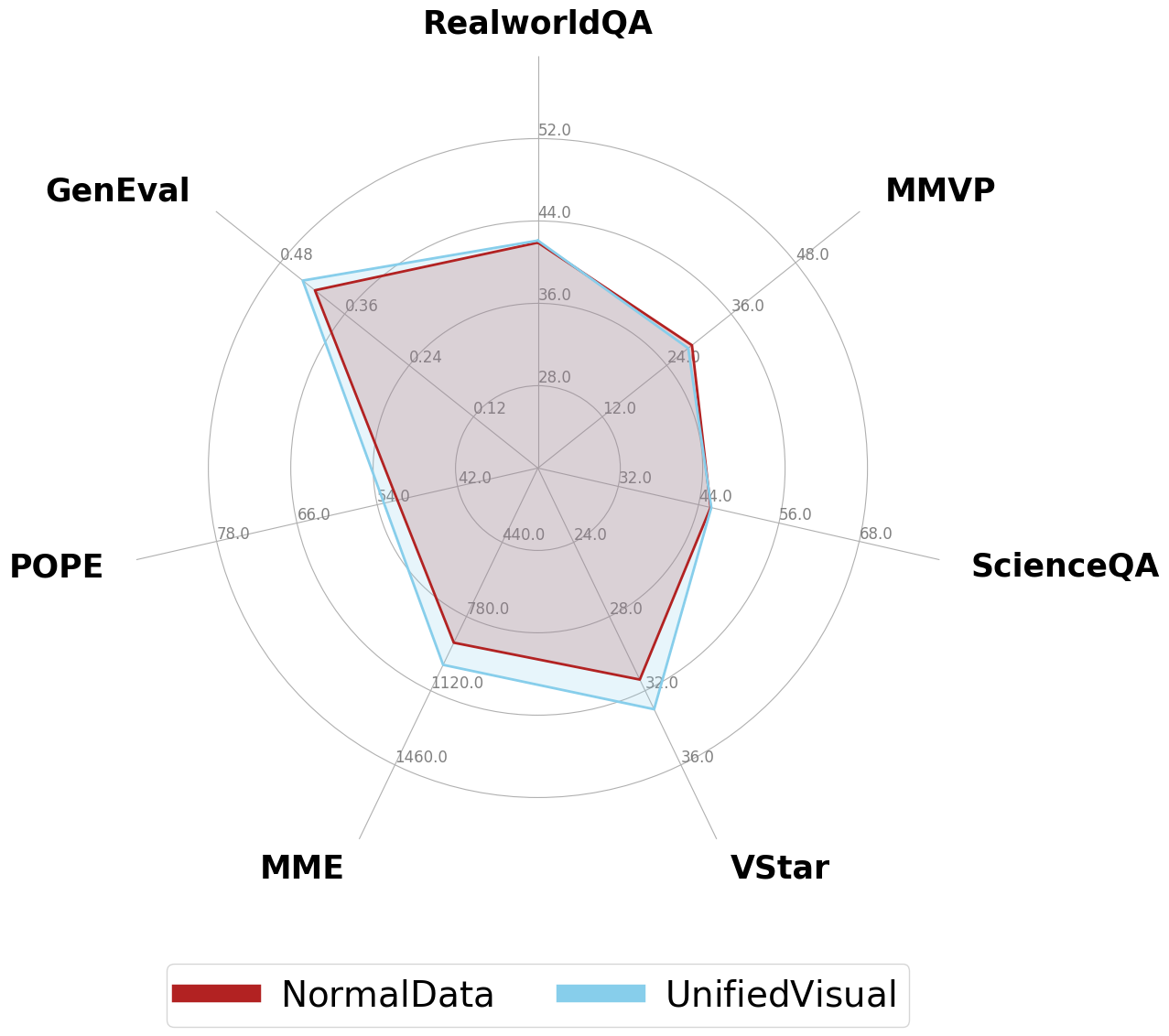}
        \caption{Emu3}
    \end{subfigure}
    \caption{Results of the multimodal understanding and generation evaluation for Anole, Liquid, and Emu3 models. Each model is trained on UnifiedVisual-240K and NormalData. The training datasets and evaluation sets are consistent with the settings described in Section \ref{sec:exp_main_res}.}
    \label{fig:results_of_various_vllms}
\end{figure*}

\subsection{Main Results}
\label{sec:exp_main_res}

\subsubsection{Multimodal Understanding}


The experimental results are presented in Table \ref{tab:mm-understanding-results}. As shown, 
compared to Anole-UnifiedVisual$_T$, which is trained solely on multimodal understanding data, Anole-NormalData incorporates additional multimodal generation data during training. However, its performance is notably worse than Anole-UnifiedVisual$_T$. This observation aligns with findings from prior research \cite{wang2024emu3}, which indicate that directly including multimodal generation data can conflict with the training objectives of multimodal understanding tasks, leading to a decline in performance compared to training exclusively on understanding data.

In contrast, our generation data is designed not only to enhance the model's generative capabilities but also to integrate complex rationales into generation tasks.
Consequently, even Anole-UnifiedVisual$_{MM}$, which is trained exclusively on our generation data, achieves slightly better performance than Anole-NormalData. More importantly, trained on the combined understanding and generation data we constructed, Anole-UnifiedVisual's understanding capability surpasses both Anole-UnifiedVisual$_T$, which is trained on our understanding data, and Anole-UnifiedVisual$_{MM}$, which is trained on our generation data.

These results clearly demonstrate that the generation data and understanding data in UnifiedVisual-240K are mutually beneficial, jointly enhancing the multimodal understanding capability of Anole-UnifiedVisual.

\subsubsection{Multimodal Generation}

As shown in figure \ref{fig:geneval-overall-score}, when trained solely on understanding data, Anole-UnifiedVisual$_T$ exhibits significantly worse generation capabilities compared to Anole-NormalData. The performance of Anole-UnifiedVisual$_{MM}$, trained on our generation data, is also slightly worse than Anole-NormalData, which may be due to the lower image quality in our UnifiedVisual dataset compared to Laion. However, when training on both the understanding and generation data in UnifiedVisual-240K, the generation capability of Anole-UnifiedVisual surpasses that of Anole-NormalData. This demonstrates that in UnifiedVisual, multimodal understanding data and multimodal generation data indeed promote each other, jointly enhancing the model's multimodal generation capability.

We further analyzed the detailed metrics of GenEval, as shown in figure \ref{fig:geneval-detailed-score}. Compared to Anole-UnifiedVisual$_{MM}$, which was trained solely on generation data, Anole-UnifiedVisual achieves significant improvements in single/double-object generation, color, and quantity. This indicates that incorporating multimodal understanding data enhances the model's comprehension of object details, including attributes such as color and quantity, thereby improving its generation capability.

To further demonstrate the advantages of the generation data in UnifiedVisual over that in NormalData, we mixed half of the NormalData generation data with half of the UnifiedVisual generation data, while keeping the understanding data consistent, and trained a new model. The resulting model achieved further improvements in generation capabilities. 
\textbf{Compared to} Anole-UnifiedVisual, this mixed-data model benefited from the introduction of higher-quality image generation data (from Laion), leading to enhanced generation performance. This finding highlights that improving image quality can further boost model performance. Additionally, \textbf{compared to} Anole-NormalData, the introduction of more complex reasoning-based generation tasks and multimodal reasoning tasks significantly enhanced the model’s generation capabilities. This further demonstrates the effectiveness of our UnifiedVisual Framework.

\subsubsection{Text Understanding}
We used AlpacaEval to evaluate the models’ text understanding and problem-solving capabilities. As shown in figure \ref{fig:alpaca-eval}, we calculated the win rate of all models compared to Anole-NormalData. Similar to the evaluation results for multimodal understanding, Anole-NormalData performs the worst, while Anole-UnifiedVisual achieves the best results. This once again demonstrates that in UnifiedVisual-240K, generation data and reasoning data mutually promote each other, thereby enhancing the model’s (textual) understanding capability.

\section{Analysis}

\subsection{Results on Various Unified VLLMs}

In this section, we further evaluate the generalizability of our UnifiedVisual by conducting experiments on additional Unified VLLMs beyond Anole, including Emu3 \cite{wang2024emu3} and Liquid \cite{wu2024liquid}. The Hugging Face repository links for these models are provided in Appendix \ref{sec:appendix_model_links}. 

As illustrated in Figure \ref{fig:results_of_various_vllms}, models trained on UnifiedVisual-240K consistently outperform those trained on NormalData in both multimodal understanding and multimodal generation tasks. These results align perfectly with our observations in Section \ref{sec:exp_main_res}, further validating the effectiveness of UnifiedVisual. Our experiments comprehensively demonstrate that UnifiedVisual substantially enhances the mutual reinforcement between multimodal understanding and generation abilities in Unified VLLMs.

\subsection{Ablation study}
\label{sec: ablation_study}

\begin{figure}[h]
    \centering
    \begin{subfigure}{.5\linewidth}
        \centering
        \includegraphics[width=\linewidth]{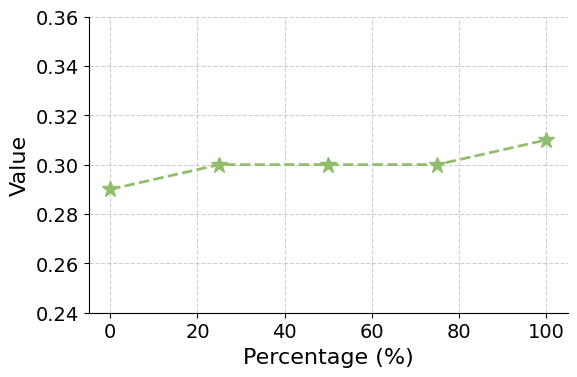}
        \caption{Generation }
        \label{fig:ablation_study_gen}
    \end{subfigure}\hfill
    \begin{subfigure}{.5\linewidth}
        \centering
        \includegraphics[width=\linewidth]{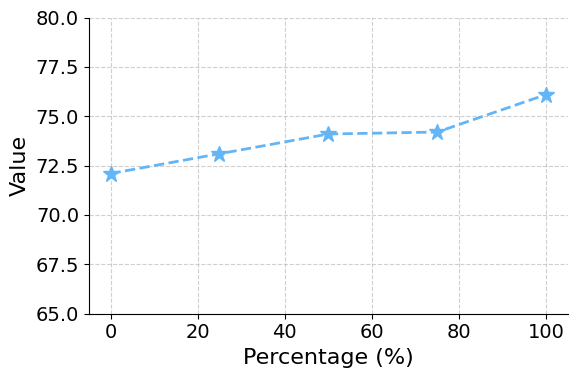}
        \caption{Understanding }
        \label{fig:ablation_study_und}
    \end{subfigure}
    \caption{\textbf{Left}: Generation performance with generation data controlled at 120k. \textbf{Right}: Understanding performance with understanding data controlled at 120k.}
    \label{fig:llava}
\end{figure}

\noindent
In this section, we further demonstrate
that training the model on UnifiedVisual-240K reveals a mutually beneficial relationship between visual understanding and generation.

\paragraph{More understanding data leads to better generation.} Building upon the findings in Section \ref{sec: main_results_mm_generation}, we conducted a controlled experiment to investigate whether more understanding data leads to better generation performance. 
In this experiment, we fixed the generation data to 120K samples and varied the amount of understanding data from 0K to 120K samples, thereby creating models with different levels of understanding. 
Figure \ref{fig:ablation_study_gen} illustrates the overall scores on GenEval, clearly demonstrating that an increase in understanding data correlates with improved generation performance.

\paragraph{More generation data leads to better understanding.} 
To explore the reverse relationship—whether more generation data enhances visual understanding—we conducted another controlled experiment. Here, we fixed the understanding data at 120K samples and vary the amount of generation data across five levels (0K, 30K, 60K, 90K, and 120K). Joint training was performed with the fixed 120K understanding samples. 
Figure \ref{fig:ablation_study_und} illustrates the  models' F1 scores on POPE, demonstrating that increasing the amount of generation data consistently improves understanding performance.
This suggests that our generation data positively impacts the model's ability to perform understanding tasks.

\paragraph{Summary.} 
our experiments confirm that, in UnifiedVisual-240K, generation and understanding data are mutually beneficial. Generation data enhances the model's multimodal understanding, while understanding data improves its generation capabilities. Additionally, we observe that the performance curves in both experiments have not yet converged. This indicates that, by following our data construction process, further scaling of the dataset could lead to even greater performance gains. Moving forward, we plan to expand the dataset to train a more powerful Unified VLLM.

\subsection{Reasoning in Multimodal Generation}

After training on UnifiedVisual-240K, Anole-UnifiedVisual demonstrates its ability to effectively leverage reasoning capabilities in visual generation tasks. As illustrated in Figure \ref{fig:appendix_mm_generation_2}, the model is prompted to generate “an animal associated with having nine lives.” While Janus-Pro-7B and Emu3-Gen were trained on larger and higher-quality datasets and can produce more realistic images, they fail to infer that the target animal was a cat. In contrast, Anole-UnifiedVisual successfully deduces that the correct animal is a cat and generates an accurate image. Additional examples are provided in Appendix \ref{sec: appendix_additional_results}. These results indicate that UnifiedVisual-240K can be used to train models to learn reasoning in multimodal generation.

\section{Conclusion}

In this paper, we propose a novel dataset construction framework, UnifiedVisual, and introduce UnifiedVisual-240K, a high-quality dataset designed to enhance the synergy between multimodal understanding and generation. Experimental results show that Anole-UnifiedVisual, trained on UnifiedVisual-240K, consistently outperforms models trained on existing datasets and demonstrates significant mutual enhancement between understanding and generation, fully validating the effectiveness of the UnifiedVisual framework.

\section*{Limitations}
In this paper, we propose a novel dataset construction framework, UnifiedVisual, and introduce a high-quality dataset, UnifiedVisual-240K. Through comprehensive experiments, we demonstrate the effectiveness of the dataset. While the current dataset is sufficient to support the experiments and conclusions presented in this paper, it remains relatively small compared to the training datasets used by other open-source models. As demonstrated in Section \ref{sec: ablation_study}, increasing the amount of training data can further enhance model performance. In the future, we plan to leverage the UnifiedVisual framework to construct larger-scale datasets, aiming to further unlock the potential of Unified VLLM.

\section*{Acknowledgements}
This work was supported by the National Natural Science Foundation of China (No. U24B20181) and Shanghai Pilot Program for Basic Research - Fudan University 21TQ1400100 (22TQ018).

\bibliography{custom}

\appendix

\clearpage

\section{Additional Experimental Setup}
\label{sec:appendix_exp_setup}

\paragraph{Training Procedure} Since both the input and output may simultaneously contain text and image content, markers [BOI] and [EOI] are added before and after the visual tokens generated from the discretization of each image. 
With visual signals fully converted into discrete tokens, we use the standard cross-entropy loss to train the model on the next-token prediction task.
Particularly, to mitigate conflicts between visual and text generation during training, we compute the loss only for text tokens when predicting text, ignoring the logits of multimodal tokens. Similarly, during visual generation, we compute the loss only for visual tokens.

\paragraph{Inference}
During inference, our model employs the next-token prediction approach. When generating text tokens, the model considers only text tokens. Once [BOI] is predicted, it signals the generation of an image. At this stage, the model focuses exclusively on predicting visual tokens until the image generation is complete.

\section{More Experimental Results}
\subsection{Detailed GenEval Score}
\begin{figure}[t]
  \centering
  \includegraphics[width=.9\columnwidth]{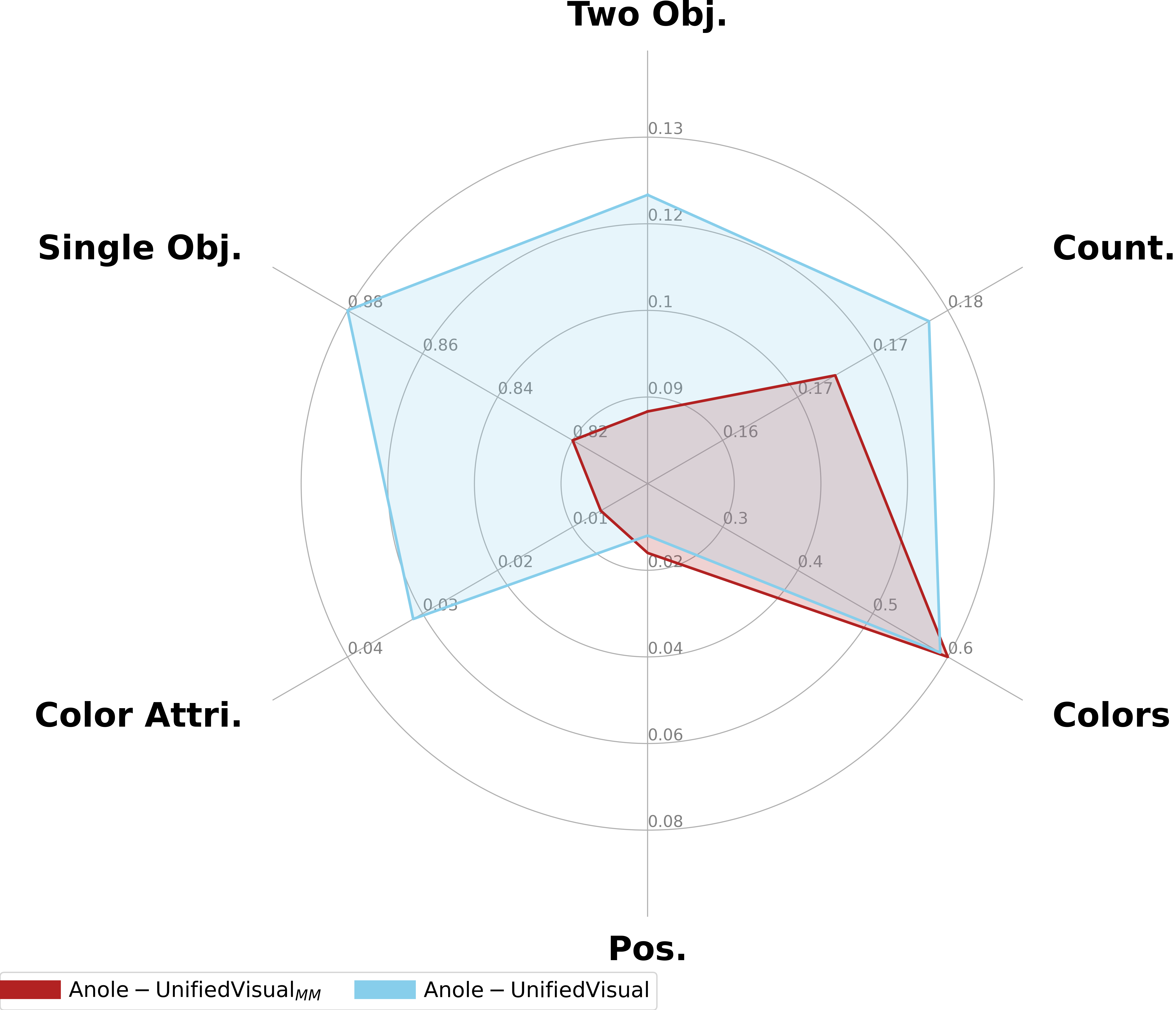}
  \caption{GenEval scores across distinct dimensions.}
  \label{fig:geneval-detailed-score}
\end{figure}

We further analyzed the detailed metrics of GenEval, as shown in figure \ref{fig:geneval-detailed-score}. Compared to Anole-UnifiedVisual$_{MM}$, which was trained solely on generation data, Anole-UnifiedVisual achieves significant improvements in single/double-object generation, color, and quantity. This indicates that incorporating multimodal understanding data enhances the model's comprehension of object details, including attributes such as color and quantity, thereby improving its generation capability.

\subsection{Evaluation on AlpacaEval}
\begin{figure}[t]
  \centering
  \includegraphics[width=.9\columnwidth]{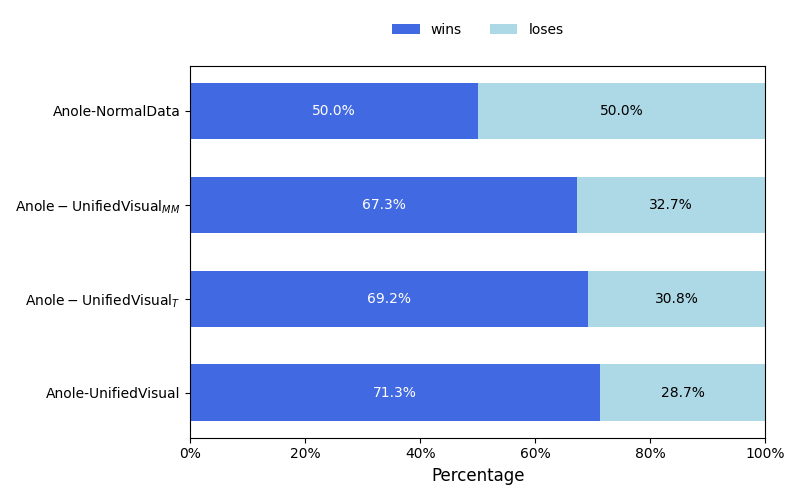}
  \caption{Evaluation on AlpacaEval.}
  \label{fig:alpaca-eval}
\end{figure}

We used AlpacaEval to evaluate the models’ text understanding and problem-solving capabilities. As shown in figure \ref{fig:alpaca-eval}, we calculated the win rate of all models compared to Anole-NormalData.

\subsection{Latency Experiments}
We evaluated the inference latency of several models on the RealWorldQA and POPE benchmarks. All experiments were conducted on a single NVIDIA H100 80G GPU. In our setup, models were required to first generate their reasoning process before producing the final answer. Table \ref{tab:latency} presents the average time (in seconds) required per sample for each model.
\begin{table}[!t]
\setlength\tabcolsep{1pt}
    \centering
    \scalebox{0.8}{
    \begin{tabular}{lcccc}
    \toprule
     & \textbf{Janus-Pro} & \textbf{LLaVA} & \textbf{Anole} & \textbf{Anole-UV} \\
    \hline
    RWQA & 2.12 & 1.06 & 1.51 & 2.07 \\
    POPE       & 1.08 & 0.59 & 1.50 & 2.03 \\
    \bottomrule
    \end{tabular}
    }
    \caption{Latency Experiments. The Hugging Face repository links for these models are provided in Appendix \ref{sec:appendix_model_links}. Anole-UV refers to Anole-UnifiedVisual.}
    \label{tab:latency}
\end{table}

Our results show that, compared to the baseline, our approach only incurs a marginal increase in inference latency, while achieving substantial improvements in performance. Furthermore, the inference latency of our model remains within an acceptable range relative to other models. We anticipate that with the adoption of inference acceleration frameworks such as vLLM, the latency can be further reduced.

\subsection{Multimodal Reasoning}


In UnifiedVisual, we introduce multimodal reasoning chains in MM Reasoning tasks with the fundamental goal of addressing the conflict between multimodal understanding and multimodal generation, thereby enabling these two capabilities to mutually enhance each other. During the reasoning process, multimodal reasoning drives the generation of necessary visual content, while the generated visual content, as part of the reasoning rationale, in turn facilitates better multimodal understanding. This design emulates human reasoning processes, where individuals often combine textual thinking with visual aids (such as mental sketches or imagined scenes) to collaboratively solve complex problems.

Through training on the UnifiedVisual-240K dataset, we observe that our models not only achieves synergistic improvements in both multimodal understanding and generation, \textbf{but also} exhibits strong multimodal reasoning abilities. 
For example, in Figure \ref{fig: main_1}, when asked, “Which plant has seeds on the outer surface of its fruit”. Models like Anole, Janus-Pro-7B, and Emu3-Gen rely on internal knowledge but give incorrect answers. 
In contrast, the Anole-UnifiedVisual model is capable of effectively "recalling" the appearances of different fruits and combining them to provide the correct answer.
This demonstrates that training on the UnifiedVisual-240K dataset  can also activate multimodal reasoning capabilities in models, allowing them to reason more like humans.

\clearpage
\onecolumn
\section{Dataset Construction}
\label{sec: appendix_dataset_construction}

\subsection{Data Sources and Quantities}

\begin{table*}[!th]
    \centering
    \resizebox{0.64\textwidth}{!}{
    \begin{tabular}{@{}lll@{}}
    \toprule
     & \textbf{Quantity} & \textbf{Source} \\
    \midrule
    MM Internet         & 29,399 & CoMM \cite{chen2024comm} \\      
    Image Editing       & 9,024  & MagicBrush \cite{Zhang2023MagicBrush}\\
    Image Generation    & 22,755 & OpenImages \cite{krasin2017openimages}\\
    Image Correction    & 20,000 & ShareGPT4V \cite{chen2024sharegpt4v}\\
    MM Reasoning (O)    & 21,000 & Visual-CoT \cite{shao2024visual}\\
    MM Reasoning (T)    & 7,276  & - \\ 
    MM Reasoning (MM)   & 17,761 & COCO \cite{lin2014microsoft}\\
    \bottomrule
    \end{tabular}
    }
    \caption{The quantities and sources of each type of generation data in UnifiedVisual-240K are presented. Here, "sources" refer to the raw data sources used to construct UnifiedVisual-240K.}
    \label{tab:UnifiedVisual-240K-generation}
\end{table*}

\subsection{Tools}

\begin{table*}[!th]
    \centering
    \scalebox{0.9}{
    \begin{tabular}{l | l}
    \toprule\
     \textbf{Tool} & \textbf{Link} \\
    \midrule
    GPT-4 & \url{https://openai.com/index/gpt-4} \\
    GPT-4o & \url{https://openai.com/index/gpt-4o-system-card} \\
    DALL-E-3 & \url{https://openai.com/index/dall-e-3} \\
    text-embedding-ada-002 & \url{https://openai.com/index/new-and-improved-embedding-model} \\
    clip-vit-large-patch14 & \url{https://huggingface.co/openai/clip-vit-large-patch14} \\
    stable-diffusion-3.5-large & \url{https://huggingface.co/stabilityai/stable-diffusion-3.5-large} \\
    Bing Image Search & \url{https://github.com/hellock/icrawler} \\
    Google Custom Search & \url{https://console.cloud.google.com} \\
    \bottomrule
    \end{tabular}}
    \caption{Links to the tools used for constructing UnifiedVisual-240K.}
    \label{tab:tool_links}
\end{table*}

\section{Links to Models}
\label{sec:appendix_model_links}
Table \ref{tab:model_links} provides links to the Hugging Face repositories for all models we use in this study.

\begin{table*}[!th]
    \centering
    \scalebox{1.0}{
    \begin{tabular}{l | l}
    \toprule\
     \textbf{Model} & \textbf{Hugging Face Page} \\
    \midrule
    Anole-7B & \url{https://huggingface.co/leloy/Anole-7b-v0.1-hf} \\
    Emu3-Stage1 & \url{https://huggingface.co/BAAI/Emu3-Stage1} \\
    Liquid & \url{https://huggingface.co/Junfeng5/Liquid_V1_7B} \\
    LLaVA-7B & \url{https://huggingface.co/liuhaotian/llava-v1.5-7b} \\
    Janus-Pro-7B & \url{https://huggingface.co/deepseek-ai/Janus-Pro-7B} \\
    \bottomrule
    \end{tabular}}
    \caption{Links to Hugging Face pages of all models.}
    \label{tab:model_links}
\end{table*}

\clearpage
\section{Illustrations of Dataset Construction Methods}
\label{sec: appendix_dataset_construction_illustration}

\begin{figure*}[h]
\centering
\includegraphics[width=1.0\textwidth]{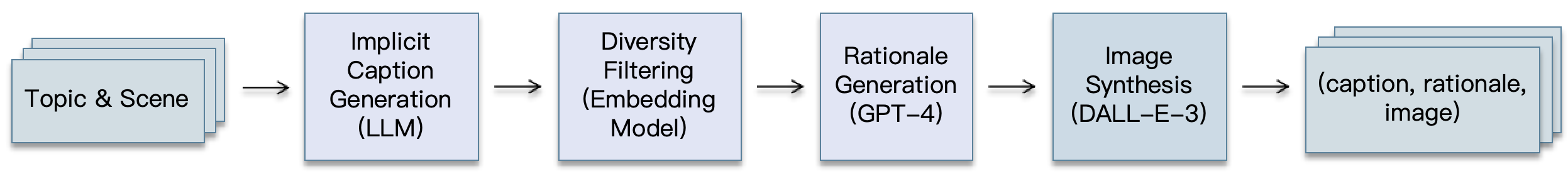}
\caption{Illustration for Topic- and Scene-Based Image Generation. (1) We propose several topics and corresponding scenes, then generate image captions that implicitly, rather than explicitly, describe the desired image content. 
(2) We use embedding models to filter similar captions, ensuring data diversity.
(3) We use GPT-4 to generate a reasoning process (rationale) explaining the content and details to be generated, (4) followed by DALL-E-3 for image synthesis.}
\end{figure*}
\begin{figure*}[h]
\centering
\includegraphics[width=1.0\textwidth]{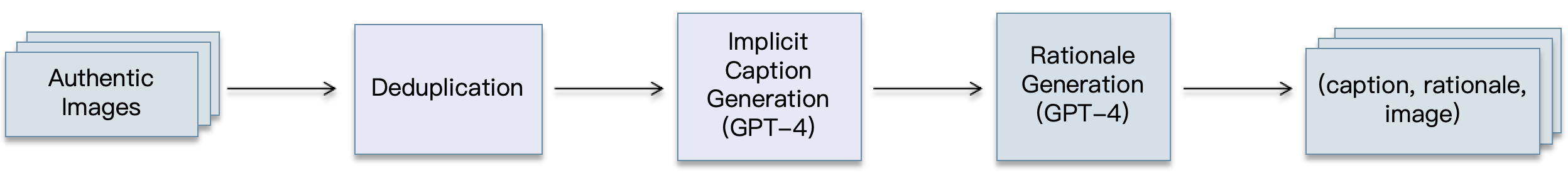}
\caption{Illustration for Category- and Image-Based. (1) We collect a diverse set of authentic images, (2) removing duplicates. (3) Based on the images and their categories, we use GPT-4 to generate instructions that describe image characteristics and related information, again focusing on implicit rather than explicit descriptions. (4) We then use GPT-4 to generate a detailed rationale based on the caption and category information, outlining the logic behind the desired image. The final data point consists of the caption, rationale, and the original image.}
\end{figure*}

\begin{figure*}[h]
\centering
\includegraphics[width=1.0\textwidth]{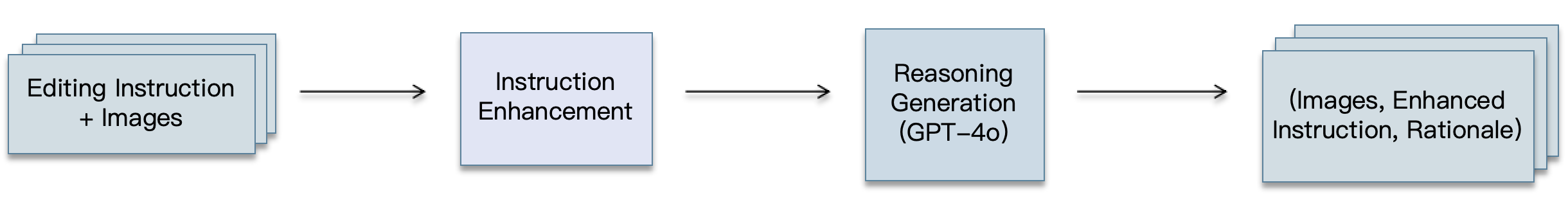}
\caption{Illustration for Image Editing. (1) We transform simple editing instructions into more nuanced prompts that necessitate deeper understanding and planning. (2) We use GPT-4o to analyze these enhanced instructions and generate reasoning rationales outlining the editing objectives and intended outcomes.}
\end{figure*}

\begin{figure*}[h]
\centering
\includegraphics[width=1.0\textwidth]{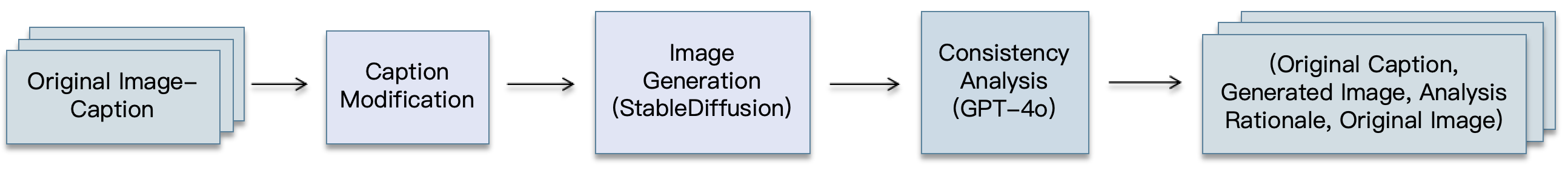}
\caption{Illustration for Image Correction. (1) We modify existing image captions to create descriptions that maintain the core theme while introducing controlled variations in specific visual elements. (2) We utilize StableDiffusion to generate images containing intentional discrepancies based on these modified descriptions. (3) We employ GPT-4o to systematically analyze the generated images against the original descriptions, automatically identifying inconsistencies and providing detailed modification rationales. The final data point includes the original caption, the generated image, the analysis rationale, and the original image.}
\end{figure*}

\begin{figure*}[h]
\centering
\includegraphics[width=1.0\textwidth]{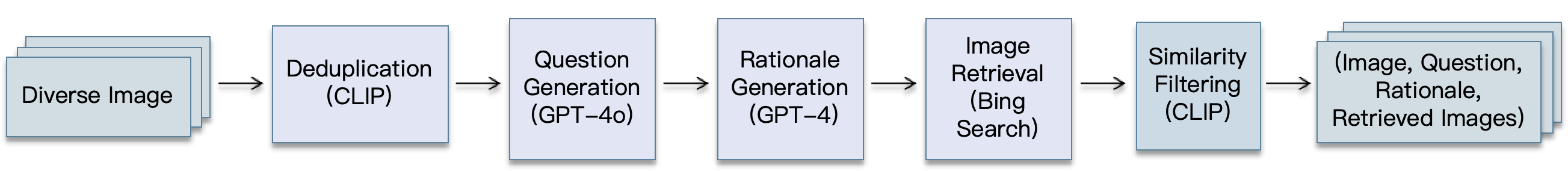}
\caption{Illustration for MM Reasoning (MM). (1) We collect a diverse set of images and use the CLIP model to remove duplicates.
(2) GPT-4o is employed to generate reasoning questions based on the collected images. These questions are designed to \textbf{require reasoning processes that integrate both visual and textual content}. Questions that fail to meet this criterion are \textbf{discarded}.
(3) The input image and the generated question are provided to GPT-4o, which produces a rationale. When necessary, textual descriptions are used in place of images.
(4) The textual descriptions from step 3 are rewritten into \textbf{keywords} using GPT-4. These keywords are then used to retrieve images from tools like Bing Search, ensuring stylistic consistency across the images used in the questions and rationales.
(5) CLIP similarity scores are computed between the descriptions generated in step 3 and the retrieved images. Only the images with the highest similarity scores are retained.}
\end{figure*}

\begin{figure*}[h]
\centering
\includegraphics[width=1.0\textwidth]{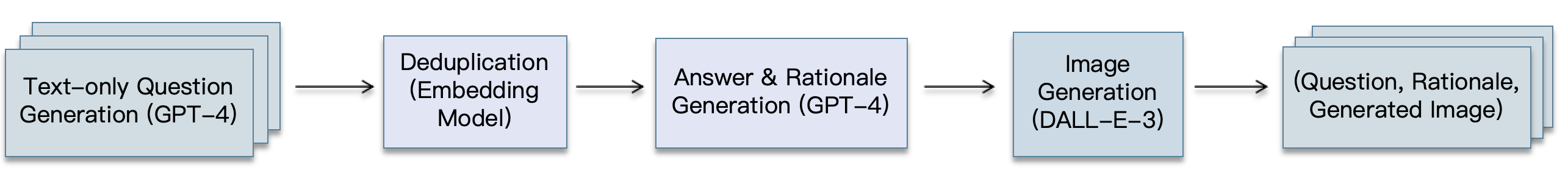}
\caption{Illustration for MM Reasoning (T). (1) GPT-4 is used to generate text-only questions that \textbf{require reasoning aided by generated images}.
(2) The generated questions are deduplicated using embedding model to ensure diversity and uniqueness.
(3) GPT-4 is then tasked with answering these questions, providing detailed rationales while replacing image-related components with textual descriptions when necessary.
(4) DALL-E-3 is used to generate images based on the image descriptions generated in step 3.}
\end{figure*}

\begin{figure*}[h]
\centering
\includegraphics[width=1.0\textwidth]{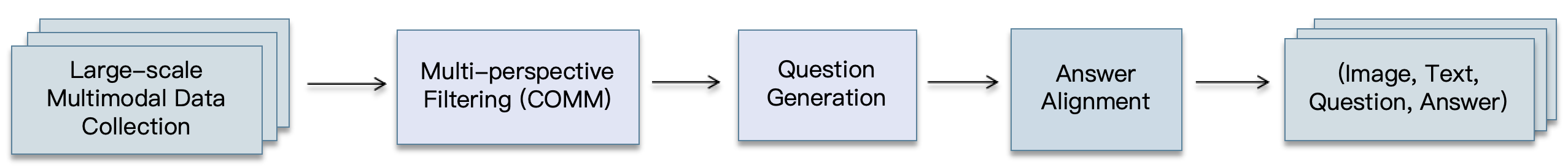}
\caption{Illustration for MM Internet. (1) We construct this dataset based on a large collection of diverse multimodal data crawled from the internet. (2) To improve data quality, we draw inspiration from \citet{chen2024comm} and design a multi-perspective filtering strategy. This strategy leverages pre-trained VLLMs to ensure coherence and semantic consistency between sentences and their associated images. (3) Furthermore, we generate questions for these multimodal data, (4) ensuring that the answers align precisely with the corresponding text-image data.}
\end{figure*}

\clearpage
\section{Prompt Template}
\label{sec:appendix_prompt_template}
\subsection{Prompt Template for Image Generation}
\label{prompt:image generation}

\begin{tcolorbox}[width=1\textwidth]
    You should generate \{number\} pairs of instruction and thought about \{topic\}. Each pair consists of:\newline\newline
    Instruction: This instruction requires generating an image. The instruction must only describe the target indirectly, without stating it explicitly (e.g., instead of "Generate an image of a panda," say, "Generate an image of the animal known for its black-and-white fur and its love for bamboo.").\newline\newline
    Thought: A detailed reasoning process that interprets the description in the instruction and deduces what should be generated. The thought should make the reasoning explicit and connect the clues to the final answer.\newline

    Examples:\newline
    \{selected examples about this topic\}\newline
    Do not include the examples in your output.
\end{tcolorbox}
\captionof{table}{Prompt template used to generate questions and rationales in Topic- and Scene-Based Generation.}
\label{tab:prompt_rationale_topic_image_generation}

\begin{tcolorbox}[width=1\textwidth]
Based on the given text, first summarize what image needs to be generated and then convert it into a format suitable for input into DALL·E 3. Just return the input for DALL·E 3, don't return anything else.\newline\newline
Text:\newline
\{thought\}\newline\newline
Input for DALL·E 3:
\end{tcolorbox}
\captionof{table}{Prompt template used to generate the DALL-E-3 input in Topic- and Scene-Based Generation.}
\label{tab:prompt_dalle_topic_image_generation}

\clearpage

\begin{tcolorbox}[width=1\textwidth]
    You will be given an object name. Your task is to:\newline\newline
    1. Create an image generation question that:\newline
       - Does not directly mention the object name\newline
       - Uses related facts, locations, or cultural references to describe it\newline
       - Requests the generation of an image \newline
    2. Provide a rationale that:\newline
       - Explains the logical connection between the facts and the object\newline
       - Ends by stating what image should be produced\newline\newline
    Output format:\newline
    \{"question": "[image generation question]","rationale": "[reasoning process and conclusion about the image to generate]"\}\newline\newline
    Examples:\newline
    object name: "the flag of the United States"\newline
    \{"question": "Show me the national flag of the country where Yellowstone National Park is located.","rationale": "Yellowstone National Park is located in the United States, so the national flag is the American flag. This means we need to create an image of the flag of the United States."
    \}\newline
    object name: "the Eiffel Tower"\newline
    \{"question": "I'd like to see an illustration of the most famous landmark in France, built as the entrance arch for the 1889 World's Fair.","rationale": "The description points to the Eiffel Tower, which was constructed for the 1889 World's Fair and stands as France's most iconic monument. The requested image should be of the Eiffel Tower."
    \}\newline
    object name: "a panda"\newline
    \{"question": "Generates an image of a black and white bear species native to the bamboo forests of central China.","rationale": "The description refers to the giant panda, which is native to China and known for eating bamboo as its main food source. The image we want is of a panda."\}\newline\newline
    Input:\newline
    object name: \{Input object name\}\newline
    Output:

\end{tcolorbox}
\captionof{table}{Prompt template used to generate questions and rationales in Category- and Image-Based Generation.}
\label{tab:prompt_rationale_category_image_generation}

\clearpage

\subsection{Prompt Template for Image Editing}

\label{prompt:image editing}

\begin{tcolorbox}[width=1\textwidth]
    I will provide you with:\newline
    An original image\newline
    An instruction for editing the image\newline
    An edited image\newline
    
    Your task is:\newline
    Based on the given before-edit image, after-edit image, and the editing instructions, analyze the differences between the two images, summarize the most notable features of the after-edit image compared to the original, and describe them in one clear and precise sentence.\newline
    It is worth noting that the Main changes include additions, deletions, and modifications, which cannot be expressed explicitly in the Output, but should be expressed implicitly.\newline
    
    Example:\newline
    \# Input:\newline
    \#\# Original image: A person sitting on a couch in a living room, looking at their phone\newline
    \#\# Editing instruction: Darken the scene, only keeping the light emitted from the phone screen\newline
    \#\# Edited image: A person sitting on a couch in a dark room, looking at their phone screen with bright light\newline
    \# Output:\newline
    It highlights the light source from the phone screen, creating a dim and focused atmosphere throughout the scene.\newline\newline
    \# Input:\newline
    \#\# Original image: \{The image before editing\}\newline
    \#\# Editing instruction: \{The original editing instruction\}\newline
    \#\# Edited image: \{The image after editing\}\newline
    \# Output:
\end{tcolorbox}
\captionof{table}{Prompt template used to generate a new editing instruction in Image Editing.}
\label{tab:prompt_instruction_image_editing}

\clearpage
\begin{tcolorbox}[width=1\textwidth]
    You are a specialized assistant for designing Image editing tasks. I will provide you with:\newline
    
    An original image\newline
    Main changes in the image after editing\newline
    An edited image\newline
    Your task is:\newline
    Convert Main changes to a question with answer about the original image that:\newline
    1. Can be a request to modify the image or a desired image\newline
    2. Must be answered with help from the edited image\newline
    3. Must be very relevant to the image and cannot be a general question that has nothing to do with the image\newline
    4. It is worth noting that the Main changes include additions, deletions, and modifications, which cannot be expressed explicitly in the question, but should be expressed implicitly.\newline

    The answer should use <image\_placeholder> to replace the edited image position in the response\newline

    \# Example:\newline
    \#\# Original image: A person sitting on a couch in a living room, looking at their phone\newline
    \#\# Edited image: A person sitting on a couch in a dark room, looking at their phone screen with bright light\newline
    \#\# Main changes: It highlights the light source from the phone screen, creating a dim and focused atmosphere throughout the scene.\newline
    
    \#\#  Question: How to highlight the light source effect of the mobile phone screen?\newline
    
    \#\# Answer: \newline
    To highlight the light source effect of the mobile phone screen, we can darken the entire scene while preserving only the light from the phone screen. This will help create contrast and emphasize the phone's light.<image\_placeholder>\newline

    \# Input:\newline
    \#\# Original image: \{The image before editing\}\newline
    \#\# Edited image: \{The image after editing\}\newline
    \#\# Main changes: \{Main changes in the image after editing\}\newline
    \# Output:
\end{tcolorbox}
\captionof{table}{Prompt template used to generate a rationale in Image Editing.}
\label{tab:prompt_rationale_image_editing}

\clearpage

\subsection{Prompt Template for Image Correction}

\label{prompt:image correction}

\begin{tcolorbox}[width=1\textwidth]
    I will give you a prompt for image generation. Please help me modify this prompt by changing or removing some key descriptive elements. The modified prompt should create an image that differs from the original in certain visual elements while maintaining the overall theme.\newline
    Prompt:\{generation prompt\}\newline
    Modified prompt:
\end{tcolorbox}
\captionof{table}{Prompt template used to generate a modified description in Image Correction.}
\label{tab:prompt_modified_image_correction}

\begin{tcolorbox}[width=1\textwidth]
    You are a professional image analysis expert.\newline
    I will provide an image generation requirement and an image generated based on that requirement. This image has some inconsistencies with the original requirements. Please analyze according to these steps:\newline
    First, carefully analyze the differences and inconsistencies between the image and the requirements.\newline
    Then, explain in detail how to make adjustments to obtain an image that fully meets the original requirements.\newline
    End with a phrase similar to "Now, let's generate a new image that fully complies with the requirements based on the above suggestions."\newline\newline
    Image generation requirement:\{generation prompt\}\newline
    Your response:
\end{tcolorbox}
\captionof{table}{Prompt template used to generate a rationale in Image Correction.}
\label{tab:prompt_rationale_image_correction}

\clearpage

\subsection{Prompt Template for MM Reasoning (MM)}

\label{prompt:MM Reasoning (MM)}

\begin{tcolorbox}[width=1\textwidth]
    [image]\newline
    Based on this image, generate a challenging analytical question that has a definitive answer. The question should:\newline\newline
    1. Require both careful observation of the image AND application of basic world knowledge\newline
    2. Require careful observation and logical reasoning to solve\newline
    3. Have a single correct answer rather than subjective interpretations\newline
    4. Be specific and precise, not vague or open-ended\newline
    5. Use world knowledge that is:\newline
        - Commonly understood and easily visualizable\newline
        - Not specialized or technical\newline
    Just provide the question without any explanation or additional information.\newline
    Question: 
\end{tcolorbox}
\captionof{table}{Prompt template used to generate a question based on an image in MM Reasoning (MM).}
\label{tab:prompt_question_MM Reasoning (MM)}

\begin{tcolorbox}[width=1\textwidth]
    [image]\newline
    You will be given an image and a question. You should analyze the image and answer the question step by step.\newline
    The rationale must be in the form of interleaved image descriptions and text. The maximum number of image descriptions in the rationale is 2.\newline
    The image descriptions and text in the rationale must complement each other to form a coherent and rigorous chain of reasoning that leads to the correct answer to the question.\newline
    The image descriptions in the response are of the form [image: description].\newline
    The image descriptions should be simple and concise enough.\newline
    The generated image descriptions cannot be close to the original image.\newline
    Just return the rationale, don't return anything else.\newline
    Question:\{question\}\newline
    Rationale:
\end{tcolorbox}
\captionof{table}{Prompt template used to generate a rationale in MM Reasoning (MM).}
\label{tab:prompt_rationale_MM Reasoning (MM)}

\clearpage

\subsection{Prompt Template for MM Reasoning (T)}

\label{prompt:MM Reasoning (T)}

\begin{tcolorbox}[width=1\textwidth]
    Please provide me with a list of \{number\} questions, options and answers about \{topic\} for Multiple Choice tasks. These questions must meet the following requirements:\newline
    Note that: The questions should have a definite answer. The answer does not change over time. Only one of the options is the correct answer. The questions and answers should not be too related to numbers.\newline
    Note that: The questions should be challenging, requiring multiple steps to answer. And the questions should be related to visual information.\newline
    Note that: The questions require a chain of thought to deduce the correct answer. The reasoning chain must be in a mixed format of text and descriptions of the images, where the descriptions of the images and text work together to form a coherent and logical chain of reasoning.\newline
    \{"question": A question generated by you, "options": 4 options in list format generated by you, "answer": The answer generated by you\}\newline\newline
    Examples:\newline
    \{selected examples about this topic\}\newline\newline
    Do not include the examples in your output.\newline
    Just provide the questions, options and answers in a jsonline format, without any explanation or additional information.
\end{tcolorbox}
\captionof{table}{Prompt template used to generate a question in MM Reasoning (T).}
\label{tab:prompt_question_MM Reasoning (T)}

\begin{tcolorbox}[width=1\textwidth]
    You will be given a multiple choice question and its correct answer. You should analyze and answer the question step by step. You need to give the rationale first and finally give the correct answer.\newline
    The rationale must be in the form of interleaved image descriptions and text.\newline 
    The image descriptions and text in the rationale must complement each other to form a coherent and rigorous chain of reasoning that leads to the correct answer to the question.\newline
    The image descriptions in the response are of the form [image: description].\newline
    Note that: The number of image descriptions in the rationale must be no more than 3.\newline
    Note that: The image descriptions of an image should contain the content of only one option.\newline
    Note that: The image descriptions should be concise and clear. \newline
    Note that: The image descriptions should be easily conveyed visually.\newline
    Just return the rationale, don't return anything else.\newline\newline
    Question: \{question\}\newline
    Options: \{options\}\newline
    Correct answer: \{answer\}\newline
    Rationale:
\end{tcolorbox}
\captionof{table}{Prompt template used to generate a rationale based on a question in MM Reasoning (T).}
\label{tab:prompt_rationale_MM Reasoning (T)}

\clearpage

\subsection{Prompt Template for Internet Multimodal Data}

\label{prompt:Internet Multimodal Data}

\begin{tcolorbox}[width=1\textwidth]
    Given the following interleaved text-image content, please generate a question for which the provided content can serve as the answer. \newline
    The images in the provided content are in the form of <image\_placeholder>. The question you generated should closely align with the logic of the provided content.\newline
    Content: \{interleaved content\}\newline
    Question:
\end{tcolorbox}
\captionof{table}{Prompt template used to generate a question based on interleaved content in Internet Multimodal Data.}
\label{tab:prompt_question_Internet Multimodal Data}

\subsection{Prompt Template for Evaluation}

\label{prompt:GPT-4 summarize}

\begin{tcolorbox}[width=1\textwidth]
    You will be provided with a question, its correct answer, and an answer to evaluate. Your task is to determine whether the given answer is correct or not. \newline
    
    \# Question:\newline
    \{question\}\newline
    
    \# Correct Answer:\newline
    \{golden answer\}\newline
    
    \# Answer to Evaluate:\newline
    \{model output\}\newline
    
    Now, determine if the answer to evaluate is correct or wrong and respond only with "Correct" or "Wrong".
\end{tcolorbox}
\captionof{table}{Prompt template used to determine whether the model’s output matches the ground truth.}
\label{tab:prompt_summarize_correct}

\begin{tcolorbox}[width=1\textwidth]
    Here is an answer to a question. This answer may be lengthy, but its final meaning is either "yes" or "no." Please carefully read and summarize the core meaning of this answer, and then determine whether its final answer is "yes" or "no." If the answer does not clearly express "yes" or "no," return "other." You must return only one word: "yes," "no," or "other.
    
    \# Question:\newline
    \{question\}\newline
    
    \# Answer:\newline
    \{model output\}

\end{tcolorbox}
\captionof{table}{Prompt template used to summarize the model’s output as 'yes' or 'no'.}
\label{tab:prompt_summarize_yes}




\clearpage

\section{Examples of Generation Data in UnifiedVisual-240K}
\label{sec:appendix_example_data}

\subsection{Example for Image Generation}
\begin{table*}[!th]
\centering
    \begin{tabular}[]{p{1.0in}p{5.0in}}
    \toprule
    \multicolumn{2}{p{5in}}{
    \textbf{Example}:}\\
    \midrule
    USER & Create an image that illustrates a sport game played in white uniforms, which lasts up to 5 days and uses a red leather ball.
    \\
    \midrule
    ASSISTANT & The description relates to Test cricket, a form of cricket that is played in matches that last up to 5 days, where players wear white uniforms, and a red leather ball is used. Hence, the image to be produced should be of a Test cricket match.\newline\newline
    \includegraphics[width=0.6\linewidth]{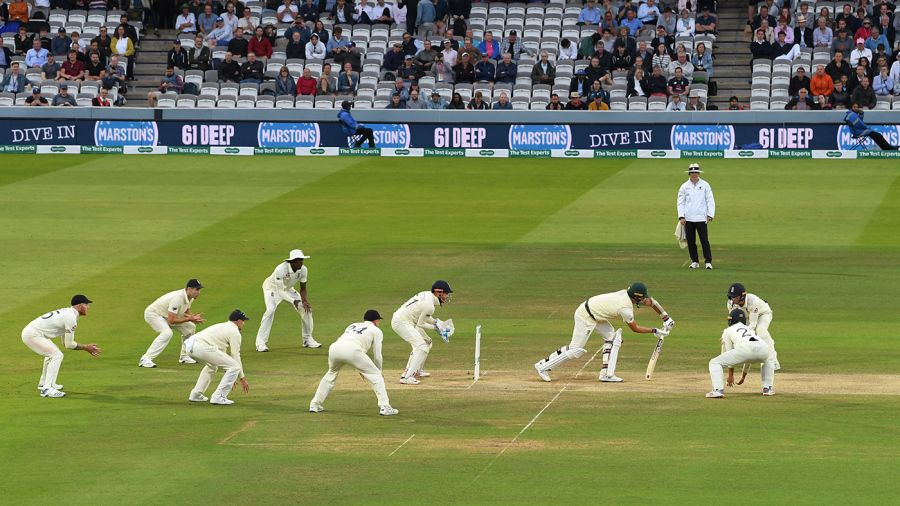}\newline
    \\

    \bottomrule
    \end{tabular}
    \caption{An example from the Image Generation subset of the UnifiedVisual-240K.}
    \label{table:example_image_generation}
\end{table*}

\clearpage

\subsection{Example for Image Editing}
\begin{table*}[!th]
\centering
    \begin{tabular}[]{p{1.0in}p{5.0in}}
    \toprule
    \multicolumn{2}{p{5in}}{
    \textbf{Example}:}\\
    \midrule
    USER & \includegraphics[width=0.6\linewidth]{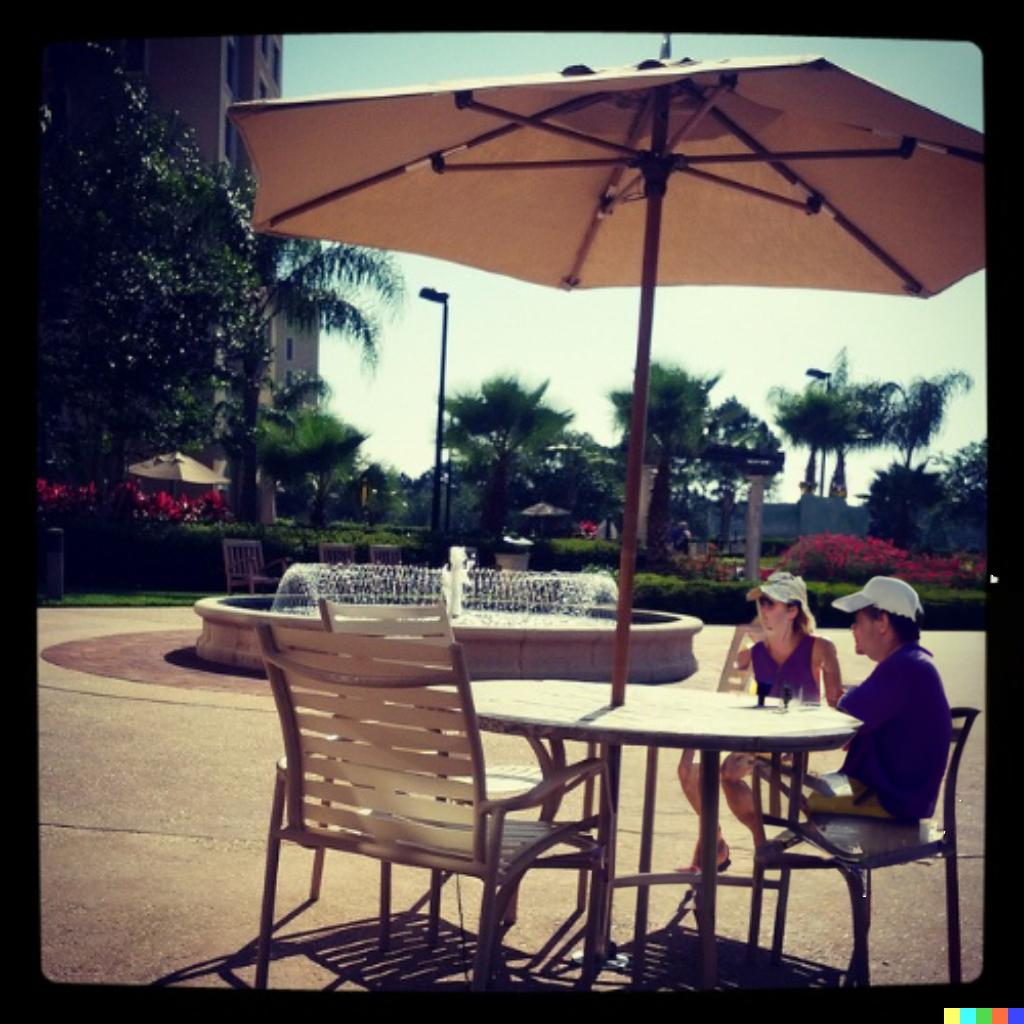}\newline
    How to achieve a serene and unoccupied patio setting focusing on just the furniture and garden scenery?
    \\
    \midrule
    ASSISTANT & To create a serene and unoccupied patio setting, we can remove any distracting elements, such as people, so the focus remains solely on the furniture and the surrounding garden scenery.\newline\newline
    \includegraphics[width=0.6\linewidth]{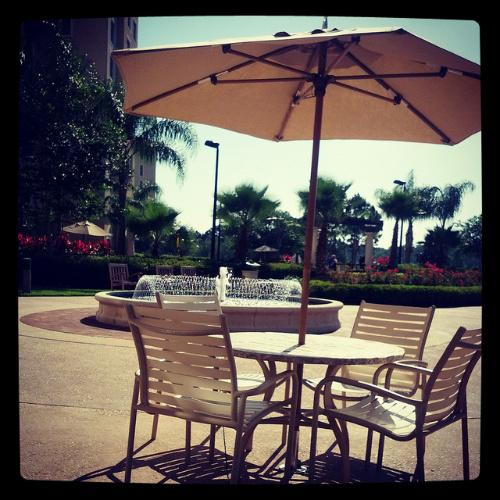}\newline
    \\

    \bottomrule
    \end{tabular}
    \caption{An example from the Image Editing subset of the UnifiedVisual-240K.}
    \label{table:example_image_editing}
\end{table*}

\clearpage

\subsection{Example for Image Correction}
\begin{table*}[!th]
\centering
    \begin{tabular}[]{p{1.0in}p{5.0in}}
    \toprule
    \multicolumn{2}{p{5in}}{
    \textbf{Example}:}\\
    \midrule
    USER & Two zebras in a grassy field: one on a dirt path facing away, with black and white stripes; the other on grass facing the camera, with brown and white stripes. Scattered rocks in the background. No other animals or text.\newline Generate an image corresponding to the text mentioned above.
    \\
    \midrule
    ASSISTANT & \includegraphics[width=0.25\linewidth]{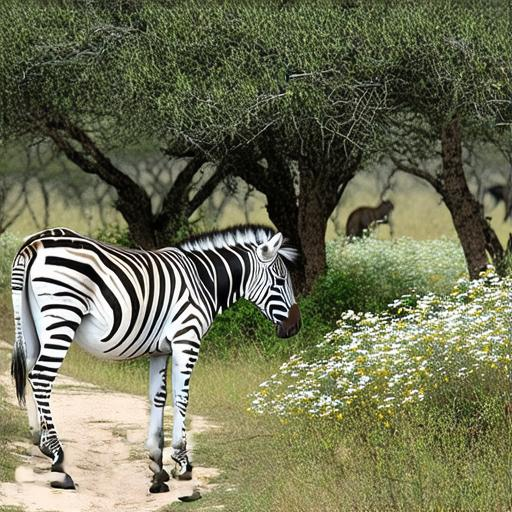}
    \\
    \midrule
    USER & The current image fails to meet the requirements. Analyze the discrepancies and generate an accurate version.
    \\
    \midrule
    ASSISTANT & 
    Upon analyzing the provided image against the specified requirements, the following inconsistencies are observed:\newline 1. \textbf{Zebra Colors and Positioning:}\newline  - The image shows only one zebra, not two. ... The image only shows one zebra on a dirt path, not fulfilling the positioning requirements.\newline 2.\textbf{Background Elements:}\newline  - The background contains additional animals, ...\newline  - There are no visible rocks scattered in the background, ...\newline \newline \textbf{Adjustments to Achieve the Requirements:}\newline- Add a Second Zebra:\newline - Introduce a second zebra with brown and white stripes into the scene, ...\newline - Correct Positioning:\newline - Retain ... on the dirt path but ensure it is facing away, as originally specified.\newline - Remove Extraneous Animals:\newline - Remove any additional animals currently visible in the background...\newline - Include Background Elements:\newline - Introduce scattered rocks into the background ...\newline Now, let's generate a new image that fully complies with the requirements based on the above suggestions.\newline\newline
    \includegraphics[width=0.3\linewidth]{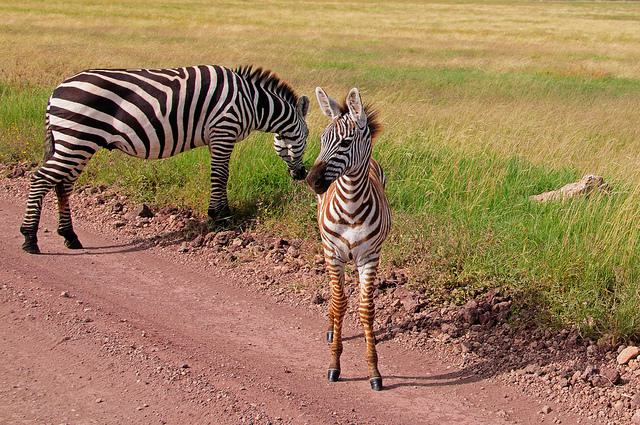}\newline
    \\
    \bottomrule
    \end{tabular}
    \caption{An example from the Image Correction subset of the UnifiedVisual-240K.}
    \label{table:example_image_correction}
\end{table*}

\clearpage

\subsection{Example for MM Reasoning (MM)}
\begin{table*}[!th]
\centering
    \begin{tabular}[]{p{1.0in}p{5.0in}}
    \toprule
    \multicolumn{2}{p{5in}}{
    \textbf{Example}:}\\
    \midrule
    USER & \includegraphics[width=0.3\linewidth]{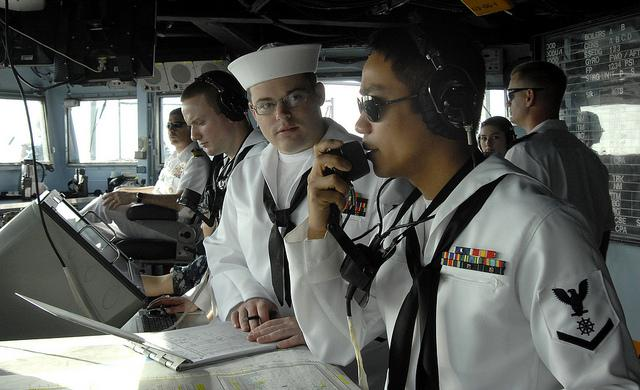}\newline
    What environment are the people in the image in?
    \\
    \midrule
    ASSISTANT & 
    The individuals are clearly part of a naval force, likely the U.S. Navy, based on their uniforms.\newline
        \includegraphics[width=0.3\linewidth]{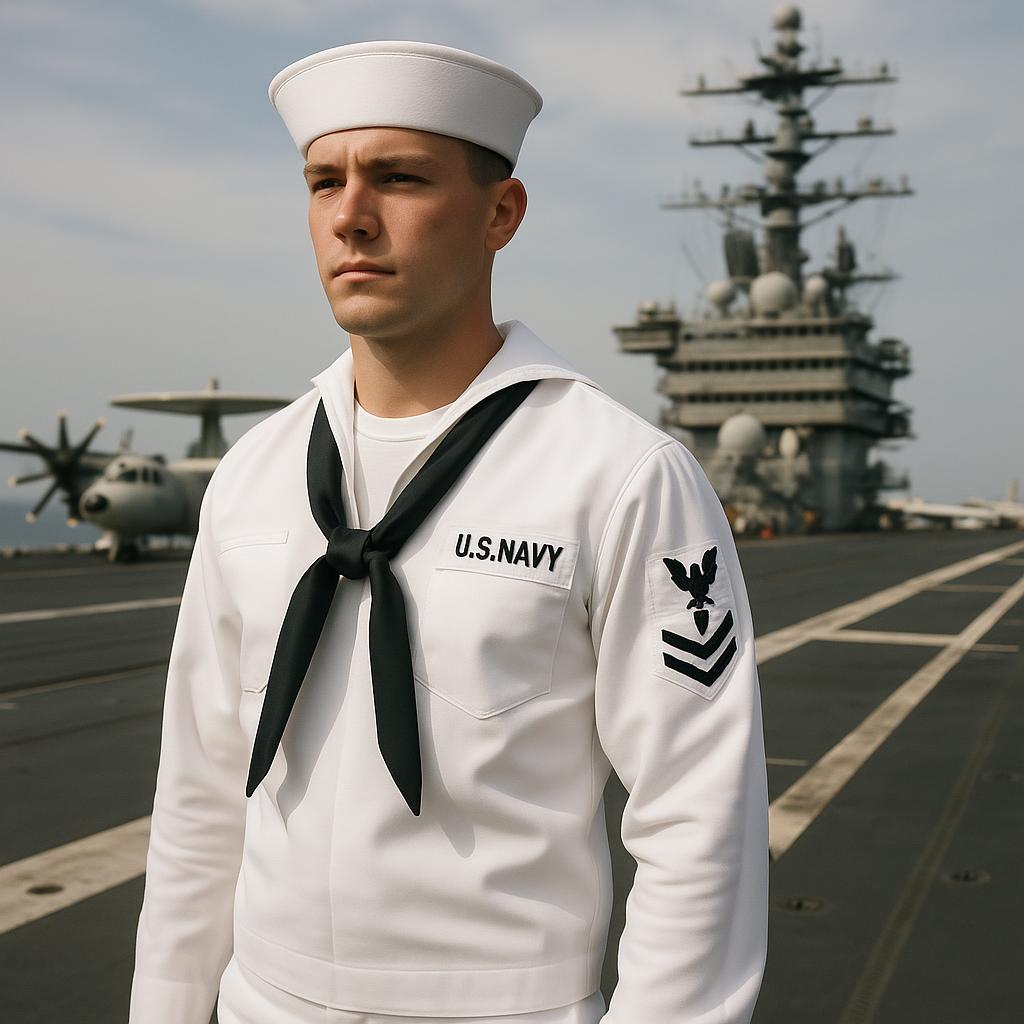}\newline\newline
    We can deduce this from their white service dress uniforms, complete with black neckerchiefs, and the distinct "dixie cup" hat worn by the man in the center. The man on the right also has rank insignia (a Boatswain's Mate rate) on his sleeve and service ribbons on his chest, further confirming their military naval affiliation.\newline
    The environment itself, with its array of control consoles, screens, charts laid out, communication headsets, and windows looking out over what appears to be water, is characteristic of the command and control center of a ship.
    \newline
        \includegraphics[width=0.3\linewidth]{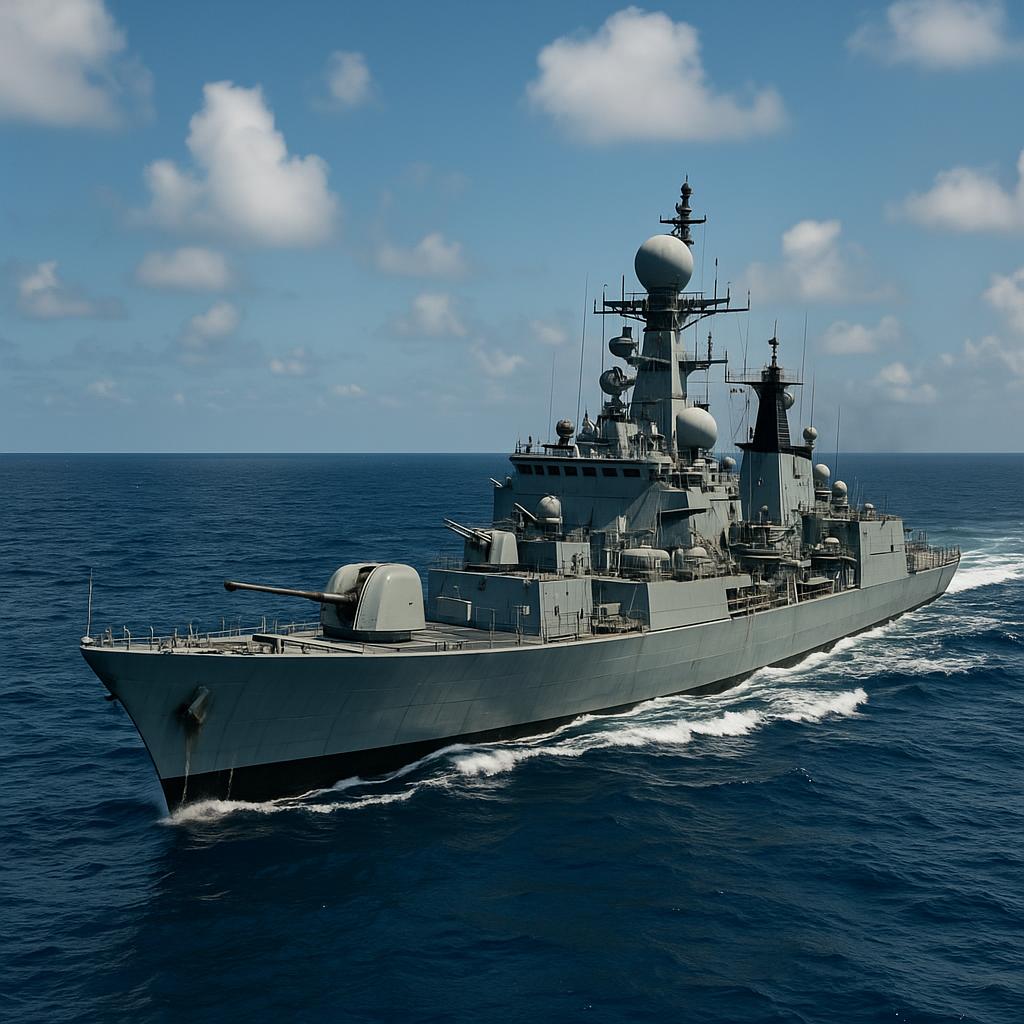}\newline\newline
    Specifically, this setup is typical of the bridge of a warship, where navigation, maneuvering, and operational command take place. The man on the right is actively using a radio handset, a common activity on a ship's bridge. The status board visible in the background right also suggests an operational environment.
    
    \\

    \bottomrule
    \end{tabular}
    \caption{An example from the MM Reasoning (MM) subset of the UnifiedVisual-240K.}
    \label{table:example_MM_Reasoning_MM}
\end{table*}

\clearpage

\subsection{Example for MM Reasoning (T)}
\begin{table*}[!th]
\centering
    \begin{tabular}[]{p{1.0in}p{5.0in}}
    \toprule
    \multicolumn{2}{p{5in}}{
    \textbf{Example}:}\\
    \midrule
    USER & Which leaf features a prominent, singular central vein running from its base to its tip, with smaller secondary veins branching off from this central vein along its path?\newline A. Maple leaf\newline B. Oak leaf\newline C. Grass blade\newline D. Ginkgo leaf
    \\
    \midrule
    ASSISTANT & The question describes a leaf with a prominent, singular central vein running from its base to its tip, with smaller secondary veins branching off from this central vein along its path. This pattern is known as pinnate venation.
    Let's look at an oak leaf:
    \newline
    \includegraphics[width=0.6\linewidth]{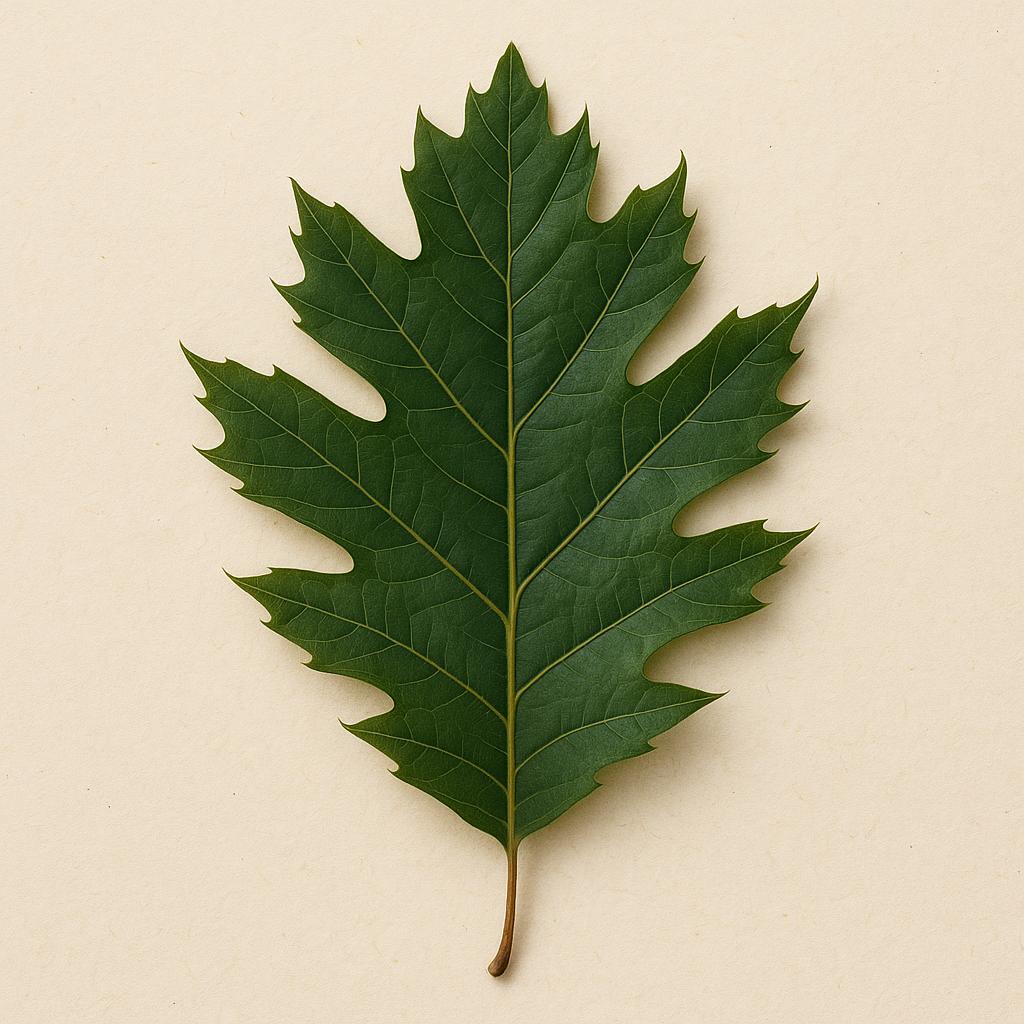}\newline\newline
    In this image of an oak leaf, you can clearly see the prominent, singular central vein (also called the midrib). It starts at the base of the leaf (where it would attach to the stem) and extends all the way to the very tip.\newline
    Along the length of this central vein, you can observe smaller secondary veins branching off to the left and right. These secondary veins then extend towards the edges and lobes of the oak leaf. This structure provides support and transports water and nutrients throughout the leaf blade.\newline
    Now let's consider why the other options are less suitable:\newline
    A. Maple leaf: Maple leaves typically have palmate venation, where several prominent veins radiate from the base, like fingers from the palm of a hand, rather than a single central one.\newline
    C. Grass blade: Grass blades exhibit parallel venation, where multiple veins run parallel to each other along the length of the blade, without a single dominant central vein from which others branch off in the described manner.\newline
    D. Ginkgo leaf: Ginkgo leaves have dichotomous venation, where veins fork repeatedly from the base, creating a fan-like pattern, again, not matching the description of a single central vein with branching secondaries.\newline
    Therefore, the correct answer is B. Oak leaf.

    \\

    \bottomrule
    \end{tabular}
    \caption{An example from the MM Reasoning (T) subset of the UnifiedVisual-240K.}
    \label{table:example_MM_Reasoning_T}
\end{table*}

\clearpage

\subsection{Example for MM Internet}
\begin{table*}[!th]
\centering
    \begin{tabular}[]{p{1.0in}p{5.0in}}
    \toprule
    \multicolumn{2}{p{5in}}{
    \textbf{Example}:}\\
    \midrule
    USER & How can you determine the sex of your chicks as they grow, and what are the key differences in appearance and behavior between male and female chicks?
    \\
    \midrule
    ASSISTANT & (1) Give it time. If you aren't in a rush to sex your chicks, you can wait until they grow enough to develop adult sex traits.  You should be able to notice adult traits within three to four weeks.   Since many of the other sexing methods require you to act quickly, this is also a good option if you get the chicks well after the initial sexing window has passed.  This is also the easiest way to determine sex for beginners.\newline
    \includegraphics[width=0.3\linewidth]{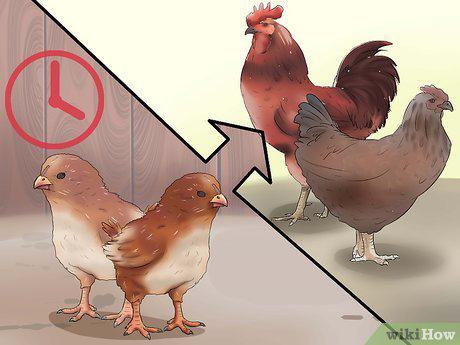}\newline\newline
    (2) Look at the appearance of the developing chick. The first trait you'll likely notice will be the comb. Male chicks develop combs within three to six weeks.  Additionally, male chicks tend to develop feathers in patchy clusters, while female chicks develop feathers more evenly.  Male chicks also tend to produce more pointed tail feathers. Female chicks, on the other hand, generally have broad, rounded feathers.  Most male chicks also develop larger feet and thicker legs than the average female chick has. This trait can be difficult to spot in young chicks but becomes increasingly more obvious as the chicks continue to age.\newline
    \includegraphics[width=0.3\linewidth]{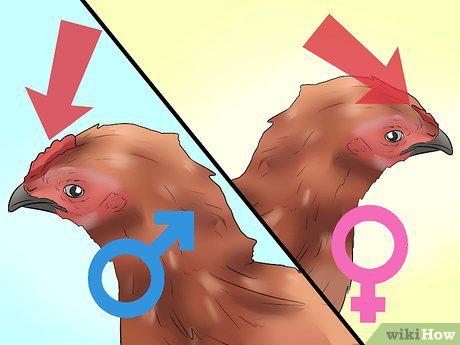}\newline\newline
    (3) Watch the chick's behavior. Male chicks demonstrate bolder, more aggressive behavior than most female chicks.  Male chicks can begin crowing within six weeks, but female chicks will only make soft clucking noises.  When startled, male chicks tend to stand their ground and give a sharp warning chirp. Females are more likely to scatter or crouch down.  Also remember that only females lay eggs. The timeline for egg laying will vary, but females can begin laying eggs within 16 to 24 weeks.\newline
    \includegraphics[width=0.3\linewidth]{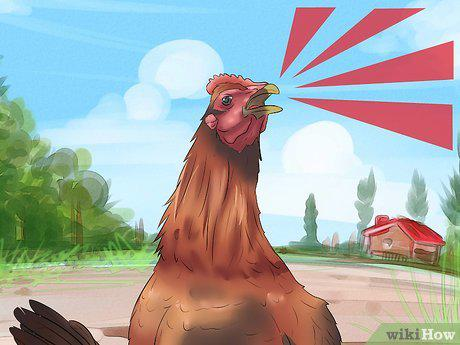}\newline\newline
    \\

    \bottomrule
    \end{tabular}
    \caption{An example from the MM Internet subset of the UnifiedVisual-240K.}
    \label{table:example_MM_Internet}
\end{table*}

\clearpage

\subsection{Example for MM Reasoning (O)}
\begin{table*}[!th]
\centering
    \begin{tabular}[]{p{1.0in}p{5.0in}}
    \toprule
    \multicolumn{2}{p{5in}}{
    \textbf{Example}:}\\
    \midrule
    USER & \includegraphics[width=0.3\linewidth]{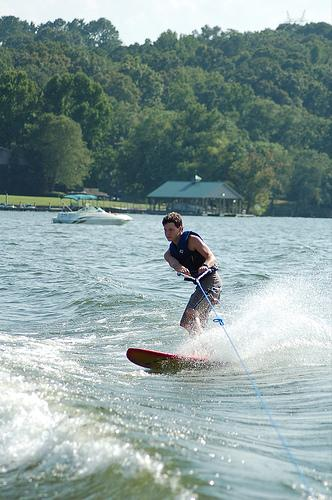}\newline
    What is the person to the right of the boat wearing?
    \\
    \midrule
    ASSISTANT & 1. Identify the object referred to as \"boat.\" 2. From the identified boat, locate the person who is positioned to the right of the boat. 3. Determine what the person located to the right of the boat is wearing. 4. State the name of the clothing item that the person is wearing.\newline The following shows the segment of the original image that is directly related to the question.\newline \includegraphics[width=0.2\linewidth]{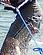}\newline\newline
    Based on the analysis and the extracted part of the original image, the final answer is determined.\newline The man is wearing shorts.
    \\

    \bottomrule
    \end{tabular}
    \caption{An example from the MM Reasoning (O) subset of the UnifiedVisual-240K.}
    \label{table:example_MM_Reasoning_O}
\end{table*}

\clearpage

\section{Additional Qualitative Results}
\label{sec: appendix_additional_results}

\begin{figure*}[h]
\centering
\includegraphics[width=1.0\textwidth]{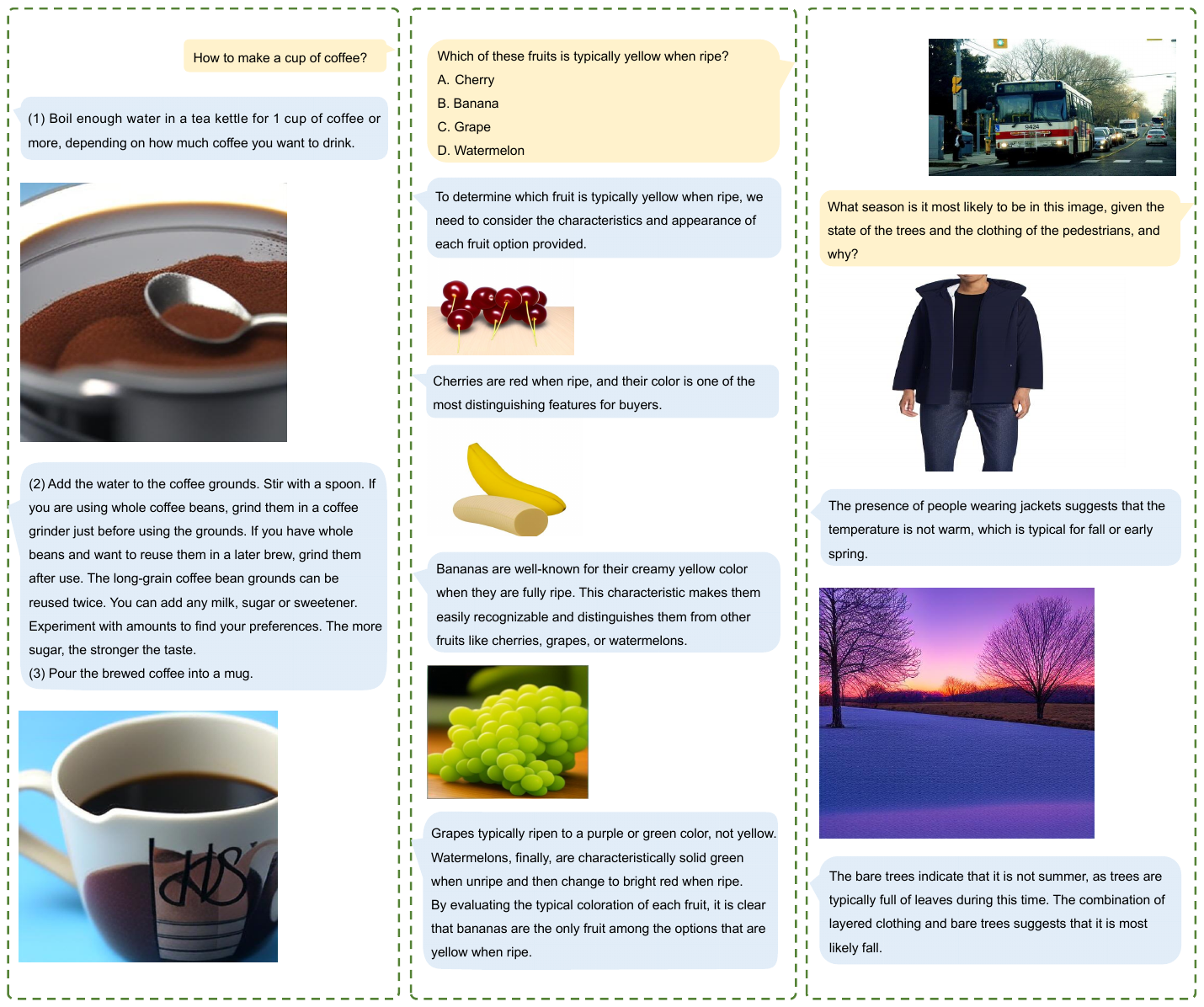}
\caption{Examples of multimodal reasoning using Anole-UnifiedVisual.}
\end{figure*}

\begin{figure*}[h]
\centering
\includegraphics[width=1.0\textwidth]{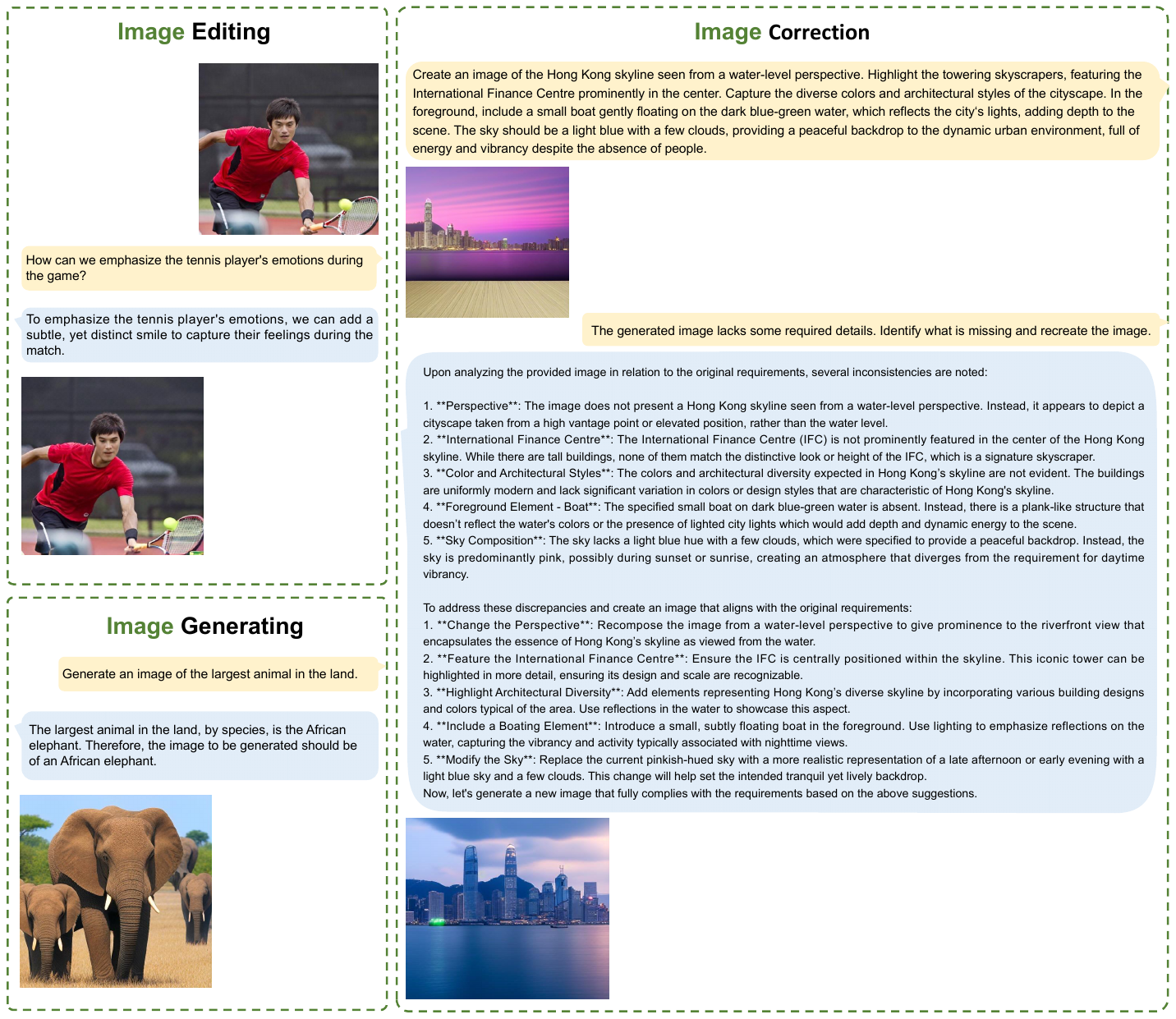}
\caption{Examples of visual generation using Anole-UnifiedVisual.}
\end{figure*}

\begin{figure*}[h]
\centering
\includegraphics[width=1.0\textwidth]{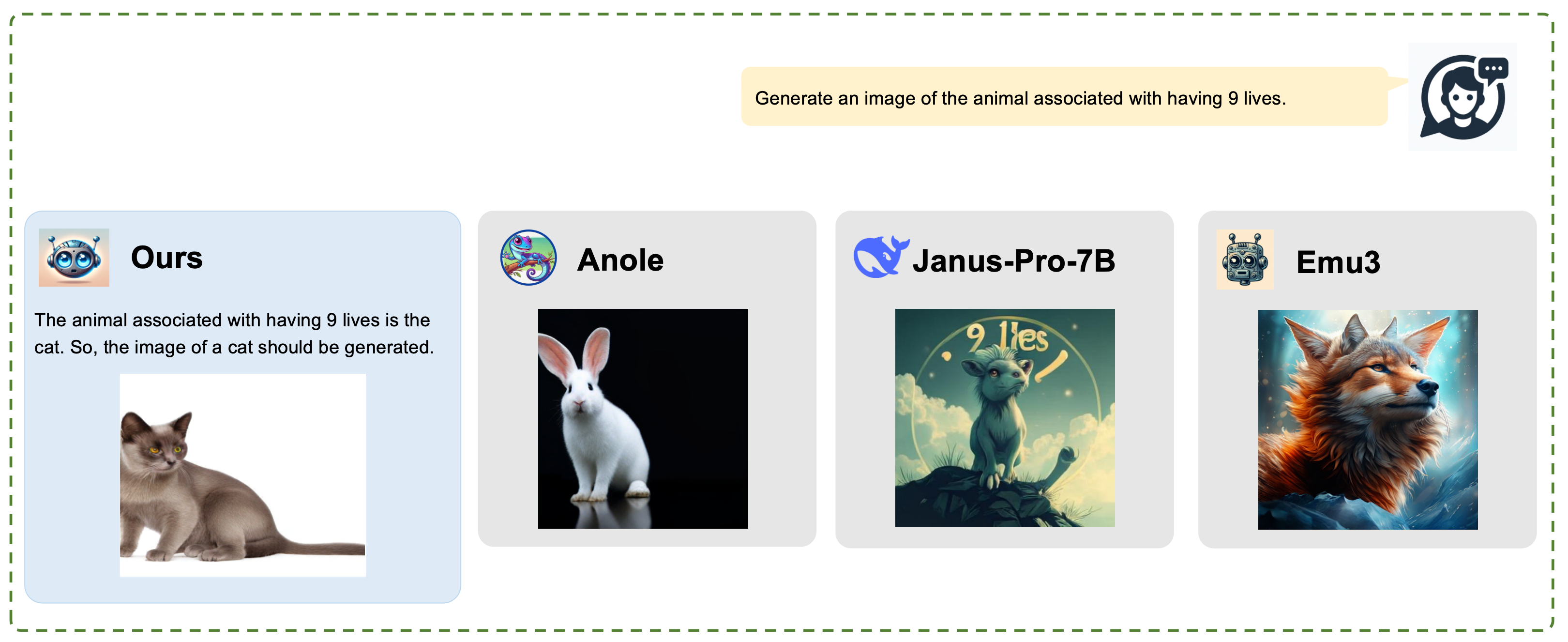}
\caption{Examples of visual generation using Anole-UnifiedVisual.}
\label{fig:appendix_mm_generation_2}
\end{figure*}

\end{document}